\documentclass[11pt]{article}

\usepackage[preprint]{acl}

\usepackage{times}
\usepackage{latexsym}

\usepackage[T1]{fontenc}

\usepackage[utf8]{inputenc}

\usepackage{microtype}

\usepackage{inconsolata}

\usepackage{graphicx}

\usepackage{xcolor}
\usepackage[most]{tcolorbox}
\usepackage{listings}
\usepackage{hyperref} 
\usepackage[capitalise]{cleveref}

\usepackage{booktabs}  
\usepackage{multirow}  
\usepackage{adjustbox} 
\usepackage{pdflscape} 
\usepackage{array}     
\usepackage{makecell}

\usepackage{graphbox}
\usepackage{wrapfig}
\usepackage{subcaption}
\usepackage{caption}
\usepackage{enumitem}

\definecolor{myred}{RGB}{200, 80, 0}  
\definecolor{mygreen}{RGB}{0, 50, 150}   

\crefname{prompt}{Prompt}{Prompts} 

\newtcolorbox{PaperBox}[2][]{
  enhanced,
  breakable,
  colback=gray!5!white,      
  colframe=blue!40!gray,    
  coltitle=black,            
  fonttitle=\bfseries,
  attach boxed title to top left={yshift=-2mm, xshift=2mm}, 
  boxed title style={colback=white, colframe=white}, 
  title={#2},
  leftrule=3pt,              
  rightrule=0.5pt, toprule=0.5pt, bottomrule=0.5pt,
  arc=2pt,
  #1
}

\newtcolorbox{ExampleBox}[1][]{
  enhanced,
  breakable,
  colback=gray!5!white,      
  colframe=blue!40!gray,    
  leftrule=3pt,              
  rightrule=0.5pt, toprule=0.5pt, bottomrule=0.5pt, 
  arc=2pt,                   
  #1
}
\newtcolorbox{PromptBox}[2][]{
  enhanced,
  breakable,
  colback=white,
  colframe=cyan!90!black,   
  title={#2},
  fonttitle=\bfseries,
  colbacktitle=cyan!90!black, 
  coltitle=white,
  attach boxed title to top center={yshift=-2mm},
  boxed title style={
    size=small,
    colframe=cyan!90!black,
    sharp corners=south
  },
  drop fuzzy shadow,
  arc=3mm,
  #1
}
\title{ConvMem: Convolutional Memory for Long-Context Reasoning}

\author{\parbox{0.92\textwidth}{\centering
  Hongming Zhang,
  Zhaozhen Gu,
  Fengshuo Bai,
  Ming Hao,
  Qingyang Zhang,
  Yuanyuan Wang,
  Shiyang Tang,
  Yanna Wang,
  Bo Xu\textsuperscript{†}}
\\[0.6ex]
  Institute of Automation, Chinese Academy of Sciences
\\[0.3ex]
  \texttt{\{hongming.zhang, boxu\}@ia.ac.cn}
}

\begin{document}
\maketitle
\begin{abstract}

While Large Language Models (LLMs) have demonstrated impressive capabilities, they often struggle with extremely long contexts due to fixed context limits. To address this, sequential approaches like MemAgent extend the effective context by reading text in segments and iteratively updating a fixed-size memory. However, this sequential paradigm suffers from high latency and requires costly reinforcement learning (RL) training, which can lead to overfitting on specific datasets. To overcome these limitations, we propose ConvMem, a training-free, highly parallelizable framework that reformulates long-context reasoning as a hierarchical convolution. Inspired by CNNs, ConvMem treats an LLM prompted with a specific query as a convolutional kernel. This kernel summarizes text segments hierarchically, shortening the reasoning path from a linear chain into a logarithmic tree. Specifically, ConvMem integrates \textit{Configurable Strides} and \textit{Skip Connections} to ensure robust evidence capture and propagation, while employing \textit{Multi-Kernel Convolution} to decompose complex queries into disentangled semantic channels. This design not only mitigates error accumulation but also enables massive parallelization across both text segments and reasoning threads. Experiments on RULER-HotpotQA and RULER-2WikiMultiHopQA demonstrate that ConvMem outperforms training-free baselines and avoids the risk of overfitting to parametric priors often observed in RL-trained models on out-of-distribution tasks.


\end{abstract}

\section{Introduction}

The capability to process extremely long contexts, ranging from digesting entire books to executing complex multi-step reasoning, has become a central objective for large language models (LLMs)~\citep{achiam2023gpt,liu2024deepseek,team2025kimi}. While advancements in positional encoding and attention mechanisms~\citep{su2024roformer,beltagy2020longformer,liu2023ring} have theoretically extended context windows to millions of tokens, practical deployment faces bottlenecks regarding inference latency and performance degradation. Specifically, standard self-attention suffers from quadratic complexity ($O(N^2)$), making it prohibitive for massive inputs. Furthermore, even models capable of processing long texts often exhibit the lost-in-the-middle phenomenon~\citep{liu2024lost}, where reasoning performance degrades as context length increases.

To address these scalability issues, recent research has pivoted towards memory-based agents, such as MemAgent~\citep{yu2025memagent}, Mem-$\alpha$~\citep{wang2025mem}, and grounded-memory agents~\citep{yang2025coarse,cui2025self}. By treating text as a sequential stream with iterative memory updates, these methods achieve linear complexity ($O(N)$). However, this paradigm introduces two critical limitations. The first is the \textit{sequential bottleneck}, where the strict temporal dependency of memory states prohibits parallelization and results in high inference latency. The second is the \textit{risk of overfitting}. To optimize memory policies, these systems rely heavily on reinforcement learning (RL)~\citep{sutton2018reinforcement,dong2020deep,zhang2020taxonomy}. Our analysis reveals that RL-trained agents tend to overfit to specific datasets, often hallucinating superficially correct answers based on pre-trained knowledge rather than faithfully reasoning over the provided context.

To overcome these challenges, we propose Convolutional Memory (ConvMem), a training-free framework designed for high-fidelity, parallelizable long-context reasoning. Drawing inspiration from convolutional neural networks (CNNs), ConvMem reformulates long-text processing by treating a frozen LLM prompted with a specific query as a semantic convolutional kernel. Unlike sequential approaches, ConvMem scans text segments concurrently at each layer. This hierarchical architecture transforms the information flow from a linear chain into a logarithmic tree structure ($O(\log N)$ depth). This topological shift not only accelerates inference via parallelization but also mitigates cumulative error propagation by shortening the reasoning path.

ConvMem incorporates three architectural mechanisms to enhance reasoning fidelity and precision:
\begin{itemize}[leftmargin=0.2in]
\item \textit{Configurable Strides for Robust Evidence Capture}: Relying on a single pass renders reasoning vulnerable to boundary truncation and model stochasticity. ConvMem introduces overlapping windows to perform multi-view scanning. This approach enables the kernel to cross-verify information across different receptive fields, enhancing the stability and recall of key evidence.
\item \textit{Skip Connections for Detail Preservation}: Hierarchical summarization inevitably compresses information, risking the loss of fine-grained details. We introduce a semantic residual mechanism where high-confidence raw evidence bypasses intermediate layers and propagates directly to the final reasoning stage, ensuring the answer is grounded in precise details.
\item \textit{Multi-Kernel Convolution for Semantic Disentanglement}: Complex tasks often involve intertwined logical threads. 
ConvMem decomposes queries into sub-questions and deploys distinct kernels to extract relevant information into separate semantic channels, preventing interference between orthogonal reasoning paths. 
\end{itemize}

We evaluate ConvMem on RULER-HotpotQA and RULER-2WikiMultiHopQA~\citep{hsieh2024ruler,yang2018hotpotqa,ho2020constructing}. Results demonstrate that ConvMem outperforms training-free baselines and exhibits superior generalization compared to RL-trained models in out-of-distribution tasks. Our contributions are threefold:
\begin{itemize}[leftmargin=0.2in]
    \item We propose ConvMem, the first training-free, CNN-inspired hierarchical framework that enables parallelizable long-context reasoning by breaking the sequential bottleneck.
    \item We empirically expose the overfitting risks of RL-based memory agents, demonstrating that their performance relies heavily on parametric memorization rather than in-context reasoning.
    \item We demonstrate that through the principled integration of configurable strides, semantic skip connections, and multi-kernel convolution, ConvMem establishes a new state-of-the-art for training-free methods, while offering superior generalization compared to RL-trained models.
\end{itemize}

\section{ConvMem for Long-Context Reasoning}

In this section, we present ConvMem, a training-free framework that adapts the architectural principles of convolutional neural networks (CNNs) to long-context reasoning. Unlike sequential memory agents that process text as a linear stream, ConvMem reformulates the reasoning process as a hierarchical, multi-channel convolutional operation. We first formulate the problem and define the semantic convolutional kernel in~\cref{sec:kernel}. We then detail the core architectural mechanisms in~\cref{sec:mechanisms}, including configurable strides, skip connections and multi-kernel convolution, that adapt visual convolution to textual reasoning. Finally, we describe the complete hierarchical parallel inference workflow in~\cref{sec:workflow}.

\subsection{Problem Formulation}
\label{sec: Preliminaries}

We consider a long-context question answering task. Let $\mathcal{D} = \{t_1, t_2, \dots, t_N\}$ denote a massive input document consisting of $N$ tokens, and $Q$ denote a user query. Our objective is to generate an answer $A$ that accurately addresses $Q$ based on the evidence contained in $\mathcal{D}$. Drawing an analogy to CNN's spatial convolution, we define key notations adapted for textual reasoning:
\begin{itemize}[leftmargin=0.2in]
    \item $W$ (\textit{Kernel Size}): The maximum context length, in tokens, scanned by a single LLM kernel.
    \item $S$ (\textit{Stride}): The step size for the scanning kernel. Setting $S < W$ creates overlapping windows, allowing each token to be scanned multiple times ($W/S$) to enhance information recall.
    \item $L$ (\textit{Depth}): The total number of layers in the hierarchical architecture.
    \item $C$ (\textit{Channels}): The number of sub-questions decomposed from $Q$.
\end{itemize}

\subsection{The Semantic Convolutional Kernel}
\label{sec:kernel}

The fundamental building block of ConvMem is the semantic convolutional kernel. Unlike standard CNN kernels that perform linear algebraic operations (dot products), our semantic kernel performs a non-linear, text-to-text transformation involving information extraction and compression.

Formally, we define the semantic kernel $\mathcal{K}$ as a function parameterized by a frozen LLM and an instruction prompt $p$. For a given layer, the kernel takes a query $q$ and a local text segment $\mathbf{x} \subset \mathcal{D}$ (analogous to the receptive field) as input. The output is a tuple consisting of a summarized text and a relevance score:
\begin{equation}
(h, r) = \mathcal{K}_{\text{LLM}}(\mathbf{x} , q ; p), 
\end{equation}
where $h$ is the condensed textual summary of $\mathbf{x}$, distilling information relevant to $q$. $r \in \{0, 1, 2\}$ is a discrete relevance score indicating the importance of the segment: $r=0$ (\textit{Irrelevant}) denotes that the segment contains no useful information; $r=1$ (\textit{Relevant}) denotes the presence of potentially useful background or context; and $r=2$ (\textit{Critical}) identifies segments containing direct evidence or the answer itself. This semantic convolution performs two functions simultaneously: information extraction (filtering noise) and dimensionality reduction (compressing token length).

To extend this local operation to the full document, analogous to CNNs scanning an entire image, we define the semantic convolution operation $\text{Conv}(\cdot)$ as follows:
\begin{equation}
\mathbf{H}, \mathbf{R} = \text{Conv}(\mathcal{D}, \mathcal{K}) = \left( \bigoplus_{i=1}^M h_i, \{r_i\}_{i=1}^M \right),
\end{equation}
$\text{where} \quad (h_i, r_i) = \mathcal{K}_{\text{LLM}}(\mathbf{x}_i, q; p)$.
This operation involves sliding the kernel $\mathcal{K}$ across all segments $\mathbf{x}_i \subset \mathcal{D}$. 
Here, $\bigoplus_{i=1}^M h_i$ denotes the sequential concatenation of local summaries $h_1, ..., h_M$, forming a global summary sequence $\mathbf{H}$.
$\mathbf{R} = [r_1, \dots, r_M]$ is the sequence of relevance scores corresponding to each $h_i$, used to identify critical segments for the semantic skip connection mechanism.

\vspace{-0.1in}
\begin{figure}[t] 
    \centering 
    \begin{subfigure}[b]{0.45\textwidth} 
        \includegraphics[width=\linewidth]{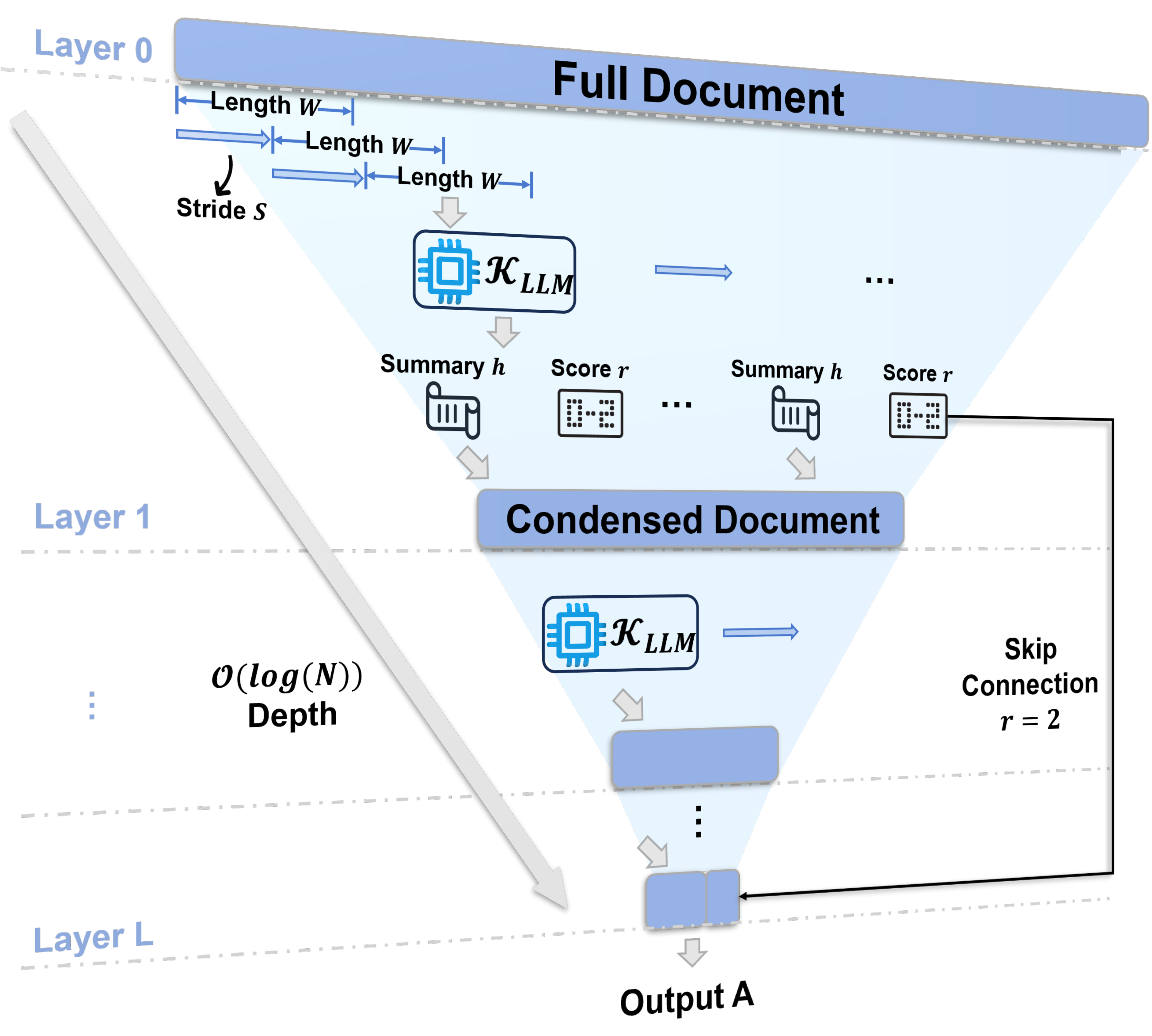}
        \vspace{-0.2in}
    \end{subfigure}
    \caption{\footnotesize Illustration of the hierarchical convolutional mechanism within a single channel. At Layer 0, the semantic kernel $\mathcal{K}$ scans raw text segments in parallel to generate condensed summaries. Subsequent layers recursively process the concatenated summaries from the previous layer. After $L$ layers, the global summary is combined with critical raw segments (skip connections) to generate the final answer $A$.
    \normalsize}
    \vspace{-0.1in}
    \label{fig:conv kernel}
\end{figure}

\subsection{CNN-Inspired Architectural Mechanisms}
\label{sec:mechanisms}

To effectively adapt the convolutional paradigm to language reasoning, ConvMem incorporates three key mechanisms: configurable strides, skip connections and multi-kernel convolution.

\subsubsection{Configurable Strides}

Standard segmentation often leads to boundary truncation, where semantic dependencies are disrupted by the rigid partitioning. We address this limitation by implementing overlapping sliding windows with stride $S < W$. The input sequence is sliced into segments where the overlap region of length $W-S$ ensures contextual continuity. Specifically, tokens at the edge of segment $i$ are re-processed near the center of segment $i+1$, preserving their contextual integrity. 
This mechanism provides two benefits: 
\begin{itemize}[leftmargin=0.2in]
\item Contextual Continuity: Overlapping windows preserve the semantic integrity that span segmentation boundaries, effectively eliminate information loss caused by arbitrary truncation.
\item Multi-View Robustness: Each token is scanned multiple times (with an over-scanning factor of $W/S$) from varying window positions. This mechanism cross-verifies information across different receptive fields, mitigating the stochasticity of the LLM kernel and resulting in a robust, ensembled understanding of the content.
\end{itemize}

\subsubsection{Skip Connections}

Hierarchical summarization entails the risk of losing fine-grained details, such as exact dates or entity names. Inspired by residual networks~\citep{he2016deep}, we introduce a semantic skip connection mechanism based on the relevance score $r$. 

During convolution at the input layer, if the kernel assigns a critical score $r=2$ to a segment $\mathbf{x}_i$, this raw segment will add a skip connection to the final layer, bypassing the summarization and is directly appended to a residual buffer $\mathcal{B}$:
\begin{equation}
\mathcal{B} \leftarrow \mathcal{B} \cup \{\mathbf{x}_i \mid \text{Score}(\mathbf{x}_i) = 2\}    
\end{equation}
This allows the final generation step to reason over high-level abstract summaries while accessing unmodified, high-fidelity evidence from the raw text.

\subsubsection{Multi-Kernel Convolution}
Complex reasoning tasks often involve tracking multiple, intertwined logical threads (e.g., ``What is the relationship between Person A and Person B?''). 
A single summary stream may conflate distinct entities or events. To handle this, we employ multi-kernel convolution to treat the text as a multi-channel signal. 

Given the user query $Q$, we first decomposes it into $C$ distinct sub-questions $\{q^{(1)}, q^{(2)}, \dots, q^{(C)}\}$. Each sub-question initializes a unique kernel $\mathcal{K}^{(c)}= \mathcal{K}_{\text{LLM}}(\cdot, q^{(c)} ; p)$. These kernels operate in parallel on the same document but focus on different semantic aspects, preventing interference between orthogonal reasoning paths. For example, $\mathcal{K}^{(1)}$ might focus on ``Person A'', while $\mathcal{K}^{(2)}$ tracks ``Person B''. This process generates $C$ distinct reasoning channels:
\begin{equation}
\mathbf{H}^{(c)} = \text{Conv}(\mathcal{D}, \mathcal{K}^{(c)})
\end{equation}
Accordingly, the global notations defined previously are extended to channel-specific versions (e.g., $\mathcal{B} \to \mathcal{B}^{(c)}$), ensuring isolation between different reasoning threads.
This mechanism ensures that the path for each sub-question remains disentangled and preserved until the final aggregation.

\subsection{Hierarchical Parallel Inference Workflow}
\label{sec:workflow}

ConvMem integrates these components into a hierarchical workflow, transforming the raw document $\mathcal{D}$ into a global answer $A$ through $L$ layers.

\paragraph{Layer 0 (Input Processing).}
The raw document $\mathcal{D}$ is sliced into $M_0$ overlapping segments $[\mathbf{x}_1^{(0)}, \mathbf{x}_2^{(0)}, \dots, \mathbf{x}_{M_0}^{(0)}]$ using window size $W$ and stride $S$. Each kernel $\mathcal{K}^{(c)}$ scans these segments in parallel to produce initial summaries and update the residual buffer $\mathcal{B}^{(c)}$.

\paragraph{Layer $l$ (Recursive Summarization).}
For hidden layers $l > 0$, we perform recursive summarization. For each channel $c$, the summaries from the previous layer are first concatenated to form an intermediate text stream $\mathbf{H}^{(l-1,c)}$.
This stream is sliced into new overlapping segments, and the semantic kernel $\mathcal{K}^{(c)}$ is applied recursively using the semantic convolution operation:
\begin{equation}
\mathbf{H}^{(l,c)} = \text{Conv}(\mathbf{H}^{(l-1,c)}, \mathcal{K}^{(c)})
\end{equation}
ConvMem achieves dimensionality reduction through semantic compression. Assuming an average compression factor $\alpha$ (the ratio of summary length to input length), the total sequence length decays exponentially: $N_l \approx N_0 \cdot \alpha^l$. As the hierarchy deepens, the effective receptive field expands exponentially, condensing the massive document into a manageable set of global representations $\mathbf{H}^{(L,c)}$.

\vspace{-0.1in}
\begin{figure}[t] 
    \centering 
    \begin{subfigure}[b]{0.45\textwidth} 
        \includegraphics[width=\linewidth]{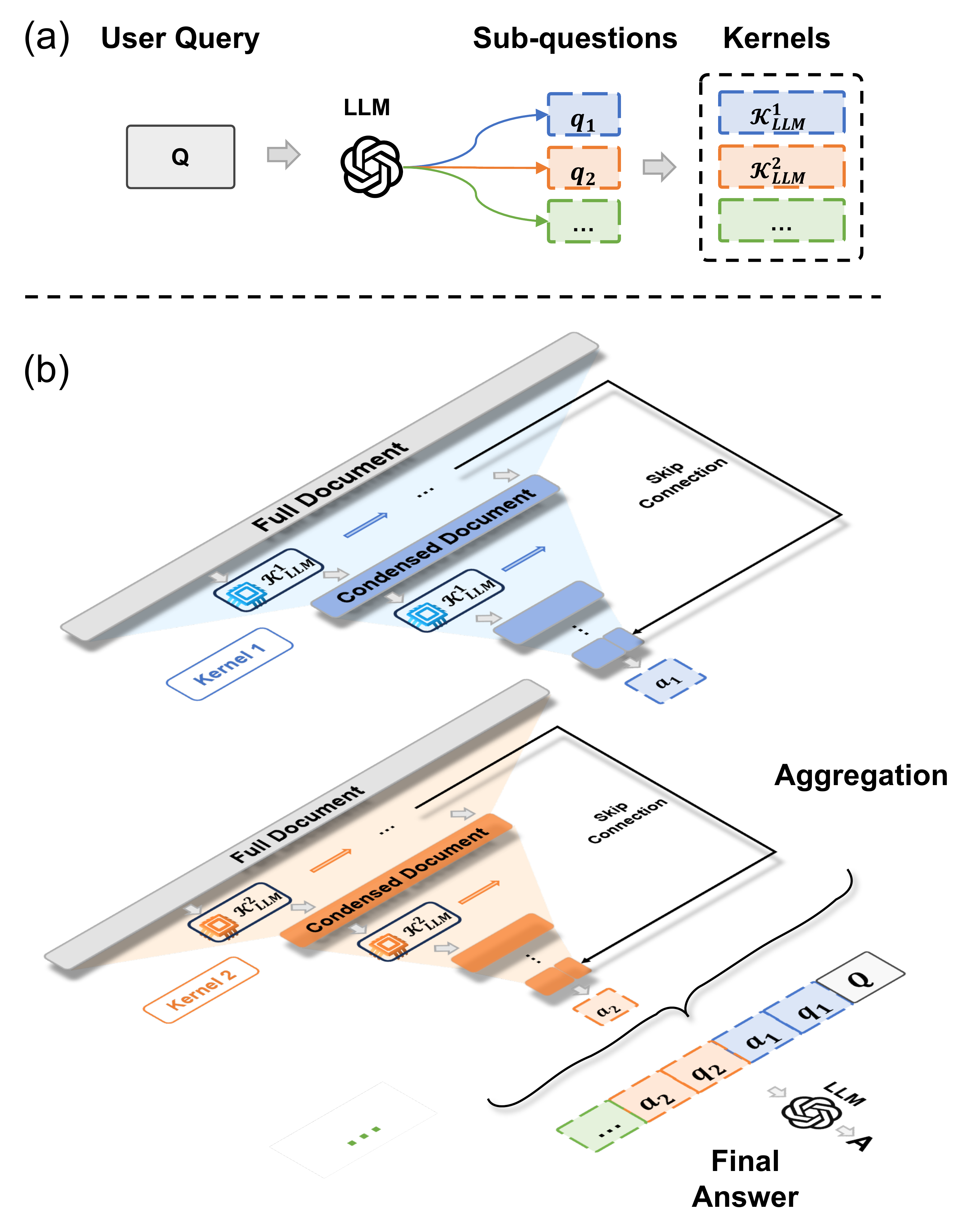}
    \end{subfigure}
    \caption{\footnotesize
    The overall workflow of ConvMem. (a) \textbf{Query Decomposition:} The complex query $Q$ is disentangled into $C$ sub-questions to isolate reasoning threads.
    (b) \textbf{Multi-Kernel Convolution:} Each sub-question drives a distinct kernel to scan the document in parallel. Through hierarchical layers, local evidence is recursively summarized and aggregated to synthesize the final global answer $A$.
    \normalsize}
    \vspace{-0.1in}
    \label{fig:overview}
\end{figure}

\paragraph{Final Aggregation.}
At the final layer $L$, the inference proceeds in two stages to generate the global answer. 
First, each channel independently generates a sub-answer $a^{(c)}$ for its sub-question $q^{(c)}$. Crucially, this step utilizes both the high-level summaries $\mathbf{H}^{(L,c)}$ and the high-fidelity raw evidence in the residual buffer $\mathcal{B}^{(c)}$ to ensure precision:
\begin{equation}
a^{(c)} = \text{LLM}(q^{(c)} \oplus \mathbf{H}^{(L,c)} \oplus \mathcal{B}^{(c)})
\end{equation}

Second, the model performs global reasoning to derive the final answer $A$. It aggregates the original query $Q$ with the sequence of resolved sub-question/sub-answer pairs:
\begin{equation}
A = \text{LLM}\left( Q \oplus \bigoplus_{c=1}^C (q^{(c)} \oplus a^{(c)}) \right)
\end{equation}
This design ensures that fine-grained details are resolved at the sub-problem level, while the final aggregation focuses on logical synthesis.

\paragraph{Parallelization.}
A critical advantage of ConvMem over sequential memory agents (e.g., MemAgent) is the decoupling of temporal dependencies. In MemAgent, step $t$ strictly depends on the memory state of step $t-1$, forcing an $O(N)$ sequential execution. In contrast, within any layer $l$ of ConvMem, the operations on all segments are independent. Furthermore, computations across different kernels $c$ are also orthogonal.

This independence enables massive parallelization. Theoretically, the inference latency is determined solely by the tree depth rather than the sequence length:
\begin{equation}
    T_{\text{latency}} \propto O(\log N)
\end{equation}
This logarithmic scaling enables ConvMem to process million-token contexts with latency comparable to processing short contexts.

\section{Experiments}
In this section, we conduct extensive experiments to investigate three key research questions: 
\textbf{Q1: Performance Efficacy.} How does ConvMem compare to standard language models and memory-enhanced agents on established benchmarks?
\textbf{Q2: Robustness and Scalability.} RL-based approaches often suffer from performance degradation when shifting to unseen data distributions. In contrast, does ConvMem demonstrate superior cross-domain consistency compared to these specialists? Furthermore, is ConvMem model-agnostic, enabling immediate performance gains when integrated different backbone LLMs without adaptation?
\textbf{Q3: Mechanism Analysis.} How do the proposed architectural components (configurable strides, skip connections, multi-kernel convolution) contribute to the system's effectiveness?

\subsection{Experimental Setup}
\paragraph{Baselines.} To ensure a comprehensive evaluation, we compare ConvMem against baselines categorized into three distinct paradigms:
\begin{itemize}[leftmargin=0.2in]
\item \textbf{Standard LLMs:} We employ \textbf{Qwen2.5-32B-Instruct}~\citep{qwen2025qwen25technicalreport} as the primary baseline. To investigate performance scaling across model sizes, we also include \textbf{Qwen2.5-7B-Instruct} and \textbf{Qwen2.5-72B-Instruct}~\citep{qwen2.5} as comparative baselines~\citep{qwen2.5}.
\item \textbf{Training-Free Memory Methods:} We evaluate retrieval-based and memory-based approaches, including \textbf{RAG-BM25}~\citep{memlab2024qwen25}, \textbf{MemAgent-W/O-RL}~\citep{yu2025memagent}, and \textbf{Mem-$\alpha$-W/O-RL}~\citep{wang2025mem}. All these methods utilize Qwen2.5-32B-Instruct as the backbone to ensure fair comparison.
\item \textbf{RL-Trained Specialists:} We benchmark against \textbf{MemAgent}~\citep{yu2025memagent} and \textbf{Mem-$\alpha$}~\citep{wang2025mem}. These models represent the RL paradigm, which optimizes sequential memory updates through extensive training.
\end{itemize}

\paragraph{Datasets.} 
We conduct evaluations on two benchmarks designed to assess both in-distribution and out-of-distribution performance:
\begin{itemize}[leftmargin=0.2in]
\item \textbf{RULER-HotpotQA (In-Distribution):} Following the MemAgent protocol, we use the RULER-HotpotQA benchmark~\citep{hsieh2024ruler,yang2018hotpotqa}. This dataset adapts multi-hop questions from HotpotQA into a long-context Needle-in-a-Haystack (NIAH) paradigm. Crucially, as MemAgent utilizes this specific data distribution for policy optimization, it serves as the in-distribution testbed.
\item \textbf{RULER-2WikiMultiHopQA (Out-of-Distribution):} To rigorously distinguish reasoning from memorization, we synthesize a novel long-context dataset based on 2WikiMultiHopQA~\citep{ho2020constructing} using the same construction protocol as RULER-HotpotQA. This dataset remains unseen during the training of MemAgent, providing a strict OOD setting to evaluate generalization.
\end{itemize}

\vspace{-0.1in}
\begin{table*}[htbp]
\centering
\caption{Main results on RULER-HotpotQA (In-Distribution) and RULER-2WikiMultiHopQA (Out-of-Distribution). We report F1 and Sub-EM scores across context lengths ranging from 28k to 896k. The best results are highlighted in bold, and the second best are underlined.}
\label{tab:main_results}

\resizebox{\linewidth}{!}{%
    \setlength{\tabcolsep}{3.0pt} 
    
    \begin{tabular}{c|c|c |cccccc |cccccc} 
    \toprule
    
    \multirow{2}{*}{\textbf{Category}} & \multirow{2}{*}{\textbf{Model}} & \multirow{2}{*}{\textbf{Metric}} 
    & \multicolumn{6}{c}{\textbf{RULER-HotpotQA}} 
    & \multicolumn{6}{c}{\textbf{2WikiMultiHopQA}} \\
    
    \cmidrule(lr){4-9} \cmidrule(lr){10-15} 
    
     & & & \textbf{28k} & \textbf{56k} & \textbf{112k} & \textbf{224k} & \textbf{448k} & \textbf{896k} 
           & \textbf{28k} & \textbf{56k} & \textbf{112k} & \textbf{224k} & \textbf{448k} & \textbf{896k} \\
    \midrule
    
    \multirow{2}{*}{\shortstack[c]{Base Model}} 
    & \multirow{2}{*}{Qwen2.5-32B-Instruct} 
     & F1      & 61.9 & 57.77 & 48.86 & 44.04 & 18.59 & 16.78 & 56.52 & 57.2 & 50.45 & 41.73 & 29.45 & 20.93 \\
    & & Sub-EM & 69.53 & 63.28 & 53.91 & 44.53 & 17.19 & 17.97 & 67.19 & 65.62 & 62.5 & 50.0 & 34.38 & 28.12 \\
    \midrule

    \multirow{6}{*}{\shortstack[c]{Training-Free \\ Methods}} 
    & \multirow{2}{*}{RAG-BM25} 
     & F1      & 45.98 & 51.11 & 47.21 & 35.86 & 29.27 & 30.67 & 34.54 & 31.17 & 24.55 & 23.19 & 21.25 & 17.34 \\
    & & Sub-EM & 46.09 & 50.00 & 48.44 & 35.94 & 30.47 & 32.81 & 57.03 & 58.59 & 47.66 & 46.88 & 39.06 & 43.75 \\
    \cmidrule{2-15}
    & \multirow{2}{*}{MemAgent-W/O-RL} 
     & F1      & 63.95 & 65.28 & \underline{62.52} & 57.98 & 55.42 & 58.73 & 60.51 & 57.42 & 45.31 & 50.6 & 47.66 & 49.65 \\
    & & Sub-EM & 67.97 & 71.09 & 65.62 & 59.38 & 56.25 & 60.16 & \underline{71.09} & 67.97 & 58.59 & 63.28 & 60.16 & 62.5 \\
    \cmidrule{2-15}
    & \multirow{2}{*}{Mem-$\alpha$-W/O-RL} 
     & F1       & 6.13 & 7.78 & 6.5 & 7.03 & 7.19 & 7.35 & 6.56 & 6.47 & 5.94 & 6.19 & 5.19 & 6.0 \\
    & & Sub-EM  & 56.25 & 53.12 & 43.75 & 46.88 & 46.88 & 49.27 & 57.38 & 59.38 & 62.5 & 53.12 & 53.12 & 50.0 \\

    \midrule
    
    \multirow{4}{*}{\shortstack[c]{RL-Trained \\ Methods}} 
    & \multirow{2}{*}{MemAgent} 
     & F1      & \textbf{75.55} & \textbf{75.20} & \textbf{75.26} & \textbf{73.54} & \textbf{73.10} & \textbf{68.80} & \underline{60.92} & \underline{60.11} & \underline{63.19} & \underline{58.82} & \underline{58.5} & \underline{58.41} \\
    & & Sub-EM & \textbf{79.69} & \textbf{79.69} & \textbf{80.47} & \textbf{77.34} & \textbf{78.91} & \textbf{74.22} & 70.31 & \underline{71.09} & \underline{72.66} & \underline{69.53} & \underline{67.97} & \underline{69.31} \\
    \cmidrule{2-15}
    & \multirow{2}{*}{Mem-$\alpha$} 
     & F1      & 5.84 & 6.81 & 7.0 & 6.09  & 1.31 & 0.69 & 1.25 & 1.13 & 0.94 & 1.03 & 0.97 & 1.09  \\
    & & Sub-EM & 34.38 & 40.62 & 40.62 & 43.75 & 43.75 & 25.0 & 59.38 & 56.25 & 50.0 & 62.5 & 50.0 & 53.12  \\
    \midrule

    \multirow{2}{*}{\shortstack[c]{Ours}} 
    & \multirow{2}{*}{\textbf{ConvMem}} 
     & F1      & \underline{67.44} & \underline{67.86} & 57.81 & \underline{63.27} & \underline{56.14} & \underline{63.09} & \textbf{72.3} & \textbf{71.25} & \textbf{67.21} & \textbf{61.96} & \textbf{61.33} & \textbf{59.06} \\
    & & Sub-EM & \underline{73.44} & \underline{72.66} & \underline{67.19} & \underline{69.53} & \underline{62.50} & \underline{69.53} & \textbf{82.81} & \textbf{82.47} & \textbf{77.34} & \textbf{71.88} & \textbf{73.31} & \textbf{70.62} \\
    \bottomrule
    \end{tabular}%
}
\end{table*}

\paragraph{Metric.} 
We report four metrics: F1 score, Exact Match (EM), Sub-EM (Sub Exact Match) and LLM-as-a-Judge ($\text{ACC}_{L}$)~\citep{rajpurkar2016squad,zheng2023judging}. F1 represents the harmonic mean of precision and recall. EM indicates whether the predicted value exactly matches the ground truth. Sub-EM measures whether the predicted value is a subset of the reference value, or vice versa. $\text{ACC}_{L}$ using LLMs as judges to evaluate the performance.
Due to space constraints, our main analysis focuses on F1 and Sub-EM, as they offer a balanced view of retrieval recall and generation flexibility. Full results across all metrics are provided in Appendix~\ref{app:full_results}.

\paragraph{Implementation Details.}
We set the kernel size $W=8000$ tokens and stride $S=1600$, implying scanning 5 times for each token. The number of channels $C$ is dynamically determined by the LLM during query decomposition. ConvMem requires no parameter updates. Prompt templates are detailed in~\cref{app:Prompt Templates}.

\subsection{Main Results and Analysis}
Table~\ref{tab:main_results} presents the performance comparison across varying context lengths. As observed, ConvMem establishes a new state-of-the-art among training-free methods and demonstrates superior robustness compared to RL-based specialists.

\paragraph{Superiority Over Training-Free Baselines.}
On both datasets, ConvMem significantly outperforms the vanilla Qwen2.5-32B-Instruct and all training-free baselines. Retrieval-based methods (RAG) struggle with multi-hop reasoning as they retrieve isolated chunks, disrupting logical dependencies. Sequential methods (MemAgent-W/O-RL) suffer from cumulative forgetting and the interference of intertwined logical threads, rendering early evidence unrecoverable.

\textbf{Case 1} illustrates this failure mode: sequential agents MemAgent-W/O-RL encounter the entity "Shirley Temple Black" early in the stream but fail to link it to the film ``Kiss and Tell'' appearing much later, as the initial memory is overwritten. In contrast, ConvMem's hierarchical aggregation preserves both facts in parallel channels, successfully retrieving the answer.

\vspace{0.5em}
\noindent\fbox{%
    \parbox{0.95\linewidth}{%
        \textbf{Case 1: Mitigating Sequential Forgetting.} \\
        \textit{Query:} What government position was held by the woman who portrayed Corliss Archer in the film Kiss and Tell? \\ 
        \textit{\textcolor{myred}{Failure Analysis} (MemAgent-W/O-RL):} The agent identifies ``Shirley Temple Black'' early but forgets her government position by the time the film ``Kiss and Tell'' appears later in the stream, due to the interference of intertwined logical threads.\\
        \textit{\textcolor{mygreen}{ConvMem Success}:} Thanks to its multi-kernel convolution and hierarchical aggregation, ConvMem retains both the entity attribute and the film relation, enabling successful multi-hop reasoning.
    }%
}
\vspace{0.5em}

\paragraph{Robustness Against RL-Induced Instability.} 
We observe that RL-based baselines suffer from distinct instability issues. Mem-$\alpha$ achieves a significantly lower F1 score compared to ConvMem. This is due to its tendency to output excessively verbose answers, which degrades precision. 

While MemAgent peaks on its training domain (RULER-HotpotQA), its performance collapses on the unseen RULER-2WikiMultiHopQA dataset. This discrepancy suggests that RL models may prioritize parametric memorization over in-context reasoning. \textbf{Case 2} provides empirical evidence of this phenomenon. The query asks to choose between ``\textbf{Lev} Yilmaz'' or ``Pamela B. Green''. However, the dataset ground truth contains a typo ``\textbf{Levni} Yilmaz''. MemAgent outputs ``\textbf{Levni} Yilmaz'', ignoring the provided context to match the memorized label. ConvMem faithfully extracts ``\textbf{Lev} Yilmaz'' from the context. Although penalized by the metric, ConvMem demonstrates superior faithfulness to the input, whereas the RL agent exhibits hallucination from priors.

\vspace{0.5em}
\noindent\fbox{%
    \parbox{0.95\linewidth}{%
        \textbf{Case 2: Correct Answer via Parametric Hallucination.} \\
        \textit{Query:} Which filmmaker was known for animation, \textbf{Lev} Yilmaz or Pamela B. Green? \\
        \textit{Dataset Label (Typo):} \textbf{Levni} Yilmaz. \\
        \textit{\textcolor{myred}{MemAgent Output}:} \textbf{Levni} Yilmaz (Matches label, contradicts context). \\
        \textit{\textcolor{mygreen}{ConvMem Output}:} \textbf{Lev} Yilmaz (Matches context, penalized as wrong).
    }%
}
\vspace{0.5em}

This case highlights a critical limitation of RL-based agents: they risk regressing into parametric retrieval systems that prioritize memorized priors over input evidence. In contrast, ConvMem, by design, functions as a faithful reasoning engine grounded in the provided context.

\begin{figure}[b]
    \centering
    \vspace{-0.1in}
    \includegraphics[width=1.0\linewidth]{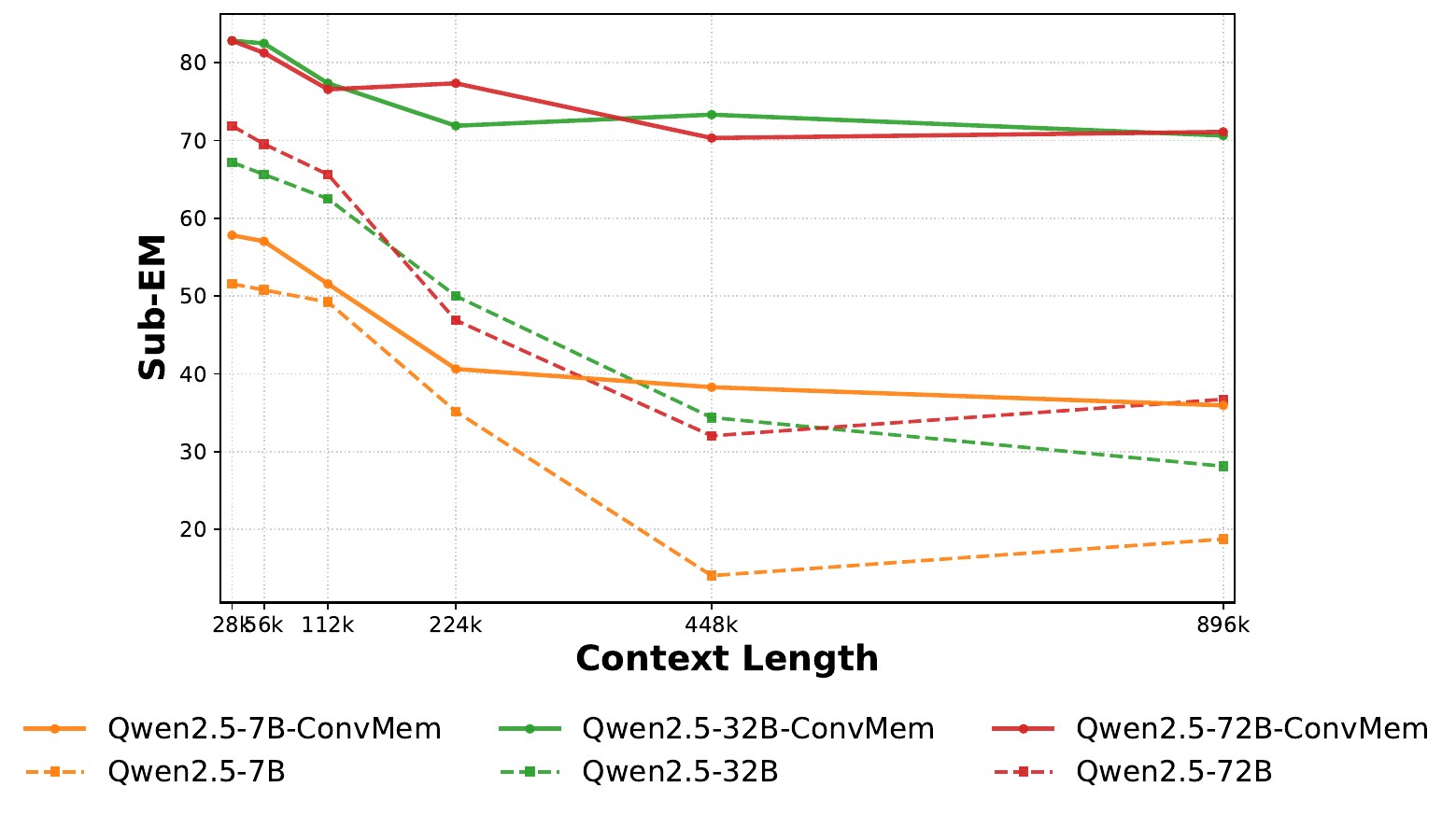}
    \vspace{-0.2in}
    \caption{ConvMem yields consistent performance improvements across diverse backbone sizes (7B-72B), validating its effective scaling capabilities.}
    \label{fig:different base model}
    \vspace{-0.1in}
\end{figure}

\vspace{-0.05in}
\paragraph{Model Agnosticism and Scalability.}
A key advantage of ConvMem is its model-agnostic nature. As shown in~\cref{fig:different base model}, applying ConvMem to backbones of varying sizes (from 7B to 72B) yields immediate performance gains. This confirms that our method effectively scales the long-context reasoning of diverse LLMs without adaptation costs.

\subsection{Ablation Study}
We conduct component-wise ablations on RULER-2WikiMultiHopQA to validate our design choices.

\begin{figure}[t]
    \centering
    \includegraphics[width=1.\linewidth]{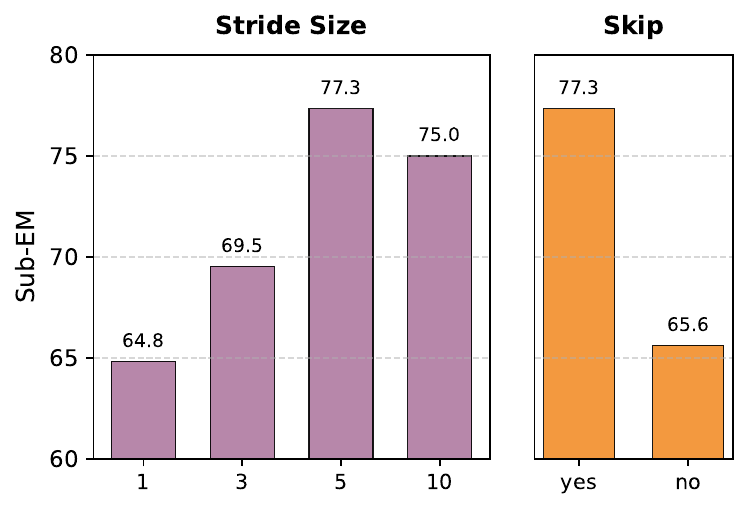}
    \vspace{-0.3in}
    \caption{\textbf{(Left)} Higher over-scanning factors ($W/S$) improve recall by mitigating segmentation boundaries, optimal at $5\times$. \textbf{(Right)} The absence of the residual path results in noticeable degradation, confirming the necessity of propagating raw high-fidelity evidence.}
    \label{fig:ablation stride skip}
    \vspace{-0.1in}
\end{figure}

\paragraph{Impact of Strides (Over-Scanning Factor).}
We investigate the effect of the \textit{over-scanning factor} ($W/S$), which determines how many times each token is processed. As shown in~\cref{fig:ablation stride skip} (Left), single-pass scanning ($S=W$) leads to a noticeable performance drop due to boundary truncation. Increasing this factor consistently improves performance, with $5\times$ over-scanning achieving the optimal balance between coverage and noise.

\paragraph{Effect of Skip Connections.}
Removing skip connections (\cref{fig:ablation stride skip}, Right) results in a degradation of fine-grained details, particularly for exact entity retrieval. This confirms that skip connections function as a semantic highway, allowing high-confidence evidence to bypass compression loss.

\vspace{-0.1in}

\begin{figure}[htbp]
    \centering
    \vspace{-0.1in}
    \includegraphics[width=1.0\linewidth]{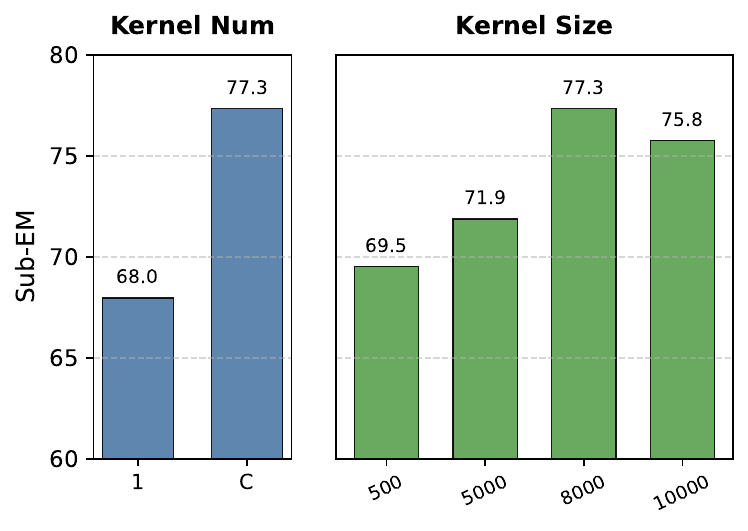}
    \vspace{-0.35in}
    \caption{\textbf{(Left)} Decomposing queries into parallel channels ($C>1$) significantly boosts performance by disentangling interference between reasoning threads. \textbf{(Right)} $W=8000$ achieves the optimal trade-off between context coherence and signal density.}
    \label{fig:ablation kernel num size}
    \vspace{-0.2in}
\end{figure}

\paragraph{Significance of Multi-Kernel Convolution.}
We compare the single-kernel against our multi-kernel convolution. Results indicate that query decomposition significantly boosts accuracy on multi-hop tasks (\cref{fig:ablation kernel num size}, Left). Single-kernel models often conflate distinct reasoning threads, whereas multi-kernel convolution successfully disentangles semantic dependencies, reducing interference.

\paragraph{Kernel Size Sensitivity.}
We test kernel sizes $W \in \{500, 5000, 8000, 10000\}$. Extremely small kernels ($500$) fragment semantic context, while overly large kernels ($10000$) dilute local signal density. $W=8000$ provides an optimal balance between context coherence and signal extraction.

\vspace{-0.1in}

\section{Related Work}

Existing long-context approaches generally fall into three categories: architectural extrapolation, memory-augmented systems, and reinforcement learning optimizations.

\subsection{Long-Context Extrapolation and Efficient Architectures}

To process inputs exceeding pre-training limits, research has focused on positional extrapolation (e.g., PI, YaRN, Ring Attention)~\citep{chen2023extending,peng2023yarn,liu2023ring} and efficiency optimizations via Linear/Sparse Attention~\citep{xiao2023efficient,katharopoulos2020transformers,child2019generating} or KV Cache compression~\citep{li2024snapkv,zhang2023h2o}. Despite these advancements, native long-context models often suffer from the lost-in-the-middle phenomenon~\citep{liu2024lost}, where reasoning performance degrades as effective context length increases. Furthermore, processing massive documents in a single pass imposes prohibitive hardware demands and latency. In contrast, ConvMem bypasses the quadratic bottleneck by decomposing the context into manageable hierarchical receptive fields, ensuring robust reasoning without modifying the underlying attention.

\subsection{Retrieval-/Memory-Augmented Agents}

To decouple context length from computational cost, recent works employ external memory systems. Retrieval-Augmented Generation (RAG)~\citep{lewis2020retrieval} reduces context by fetching top-$k$ chunks, but it often fails on multi-hop reasoning tasks due to the retrieval of isolated fragments lacking global connectivity. Alternatively, Sequential Memory Agents, such as MemGPT~\citep{packer2023memgpt}, Mem0~\citep{chhikara2025mem0}, and MIRIX~\citep{wang2025mirix}, MemAgent~\citep{yu2025memagent} treat text as a continuous stream, maintaining a persistent memory state that evolves step-by-step. While effective, these sequential methods introduce a temporal dependency ($O(N)$) that prohibits parallelization, leading to high inference latency. ConvMem addresses these limitations by adopting a tree-structured, CNN-inspired topology ($O(\log N)$), enabling massive parallelization and shortening the information propagation path to minimize error accumulation.

\vspace{-0.1in}
\subsection{Reinforcement Learning for Memory Optimization}

Acknowledging that heuristic memory updates may be suboptimal, recent works employ RL to learn memory policies, following the long tradition of exploiting memory in RL~\citep{zhang2023replay,zhang2024exploiting,zhang2024provable}. For instance, MemAgent~\citep{yu2025memagent} utilizes Multi-Conv DAPO to optimize memory overwrite decisions, while Memory-R1~\citep{yan2025memory} and Mem-$\alpha$~\citep{wang2025mem} train agents to manage complex memory structures via reward signals derived from downstream task accuracy. While RL yields performance gains on in-domain benchmarks, it introduces significant training costs and are susceptible to overfitting dataset. In contrast, ConvMem proposes a training-free paradigm. By leveraging the intrinsic instruction-following capabilities of pre-trained LLMs rather than biased reward optimization, ConvMem achieves superior robustness and true in-context reasoning.

\section{Conclusion}

In this work, we propose \textbf{ConvMem}, a training-free framework that reformulates long-context reasoning as a hierarchical, multi-channel convolutional process. By reconceptualizing LLM-based reasoning through the lens of Convolutional Neural Networks, ConvMem successfully transforms the linear dependency chain ($O(N)$) into a logarithmic tree structure ($O(\log N)$), effectively mitigating cumulative error propagation. Our empirical evaluation reveals that this principled adaptation, leveraging multi-kernel convolution, configurable strides, and skip connections, achieves competitive performance on standard benchmarks while demonstrating superior generalization on out-of-distribution tasks compared to RL-trained specialists. Furthermore, our analysis exposes the tendency of RL-based agents to overfit to dataset artifacts, suggesting that ConvMem’s training-free, context-anchored approach provides a more robust path for genuine in-context reasoning. We hope this work encourages the community to further explore architectural innovations that balance computational efficiency with reasoning fidelity.

\clearpage
\section{Limitations}
While ConvMem offers significant advantages in latency and robustness, we acknowledge two primary limitations that point towards future research directions. First, regarding token consumption vs. latency trade-off, although ConvMem achieves logarithmic latency through parallelization, the total computational cost is higher than that of linear scanning methods. Second, the system exhibits a dependency on query decomposition. The efficacy of our parallel convolution relies on the backbone model's ability to correctly disentangle complex queries into orthogonal sub-questions. If the initial decomposition is flawed, subsequent kernels may operate on incomplete premises.

\section{Ethical Considerations}
This work proposes a training-free framework for long-context reasoning. While our approach requires multiple passes over the text, increasing inference computation, it eliminates the energy consumption associated with model training. Limitations regarding potential biases stem from the underlying frozen LLMs used as kernels. All datasets used in this study are publicly available, and no private data was involved in the experiments.

\bibliography{custom}

@article{liu2024lost,
  title={Lost in the middle: How language models use long contexts},
  author={Liu, Nelson F and Lin, Kevin and Hewitt, John and Paranjape, Ashwin and Bevilacqua, Michele and Petroni, Fabio and Liang, Percy},
  journal={Transactions of the Association for Computational Linguistics},
  volume={12},
  pages={157--173},
  year={2024}
}

@article{beltagy2020longformer,
  title={Longformer: The long-document transformer},
  author={Beltagy, Iz and Peters, Matthew E and Cohan, Arman},
  journal={arXiv preprint arXiv:2004.05150},
  year={2020}
}

@article{xiao2023efficient,
  title={Efficient streaming language models with attention sinks},
  author={Xiao, Guangxuan and Tian, Yuandong and Chen, Beidi and Han, Song and Lewis, Mike},
  journal={arXiv preprint arXiv:2309.17453},
  year={2023}
}

@article{su2024roformer,
  title={Roformer: Enhanced transformer with rotary position embedding},
  author={Su, Jianlin and Ahmed, Murtadha and Lu, Yu and Pan, Shengfeng and Bo, Wen and Liu, Yunfeng},
  journal={Neurocomputing},
  volume={568},
  pages={127063},
  year={2024},
  publisher={Elsevier}
}

@inproceedings{katharopoulos2020transformers,
  title={Transformers are rnns: Fast autoregressive transformers with linear attention},
  author={Katharopoulos, Angelos and Vyas, Apoorv and Pappas, Nikolaos and Fleuret, Fran{\c{c}}ois},
  booktitle={International conference on machine learning},
  pages={5156--5165},
  year={2020},
  organization={PMLR}
}

@article{child2019generating,
  title={Generating long sequences with sparse transformers},
  author={Child, Rewon and Gray, Scott and Radford, Alec and Sutskever, Ilya},
  journal={arXiv preprint arXiv:1904.10509},
  year={2019}
}

@article{liu2023ring,
  title={Ring attention with blockwise transformers for near-infinite context},
  author={Liu, Hao and Zaharia, Matei and Abbeel, Pieter},
  journal={arXiv preprint arXiv:2310.01889},
  year={2023}
}

@article{yu2025memagent,
  title={MemAgent: Reshaping Long-Context LLM with Multi-Conv RL-based Memory Agent},
  author={Yu, Hongli and Chen, Tinghong and Feng, Jiangtao and Chen, Jiangjie and Dai, Weinan and Yu, Qiying and Zhang, Ya-Qin and Ma, Wei-Ying and Liu, Jingjing and Wang, Mingxuan and others},
  journal={arXiv preprint arXiv:2507.02259},
  year={2025}
}

@article{wang2025mem,
  title={Mem-$\{$$\backslash$alpha$\}$: Learning Memory Construction via Reinforcement Learning},
  author={Wang, Yu and Takanobu, Ryuichi and Liang, Zhiqi and Mao, Yuzhen and Hu, Yuanzhe and McAuley, Julian and Wu, Xiaojian},
  journal={arXiv preprint arXiv:2509.25911},
  year={2025}
}

@inproceedings{he2016deep,
  title={Deep residual learning for image recognition},
  author={He, Kaiming and Zhang, Xiangyu and Ren, Shaoqing and Sun, Jian},
  booktitle={Proceedings of the IEEE conference on computer vision and pattern recognition},
  pages={770--778},
  year={2016}
}

@misc{qwen2025qwen25technicalreport,
      title={Qwen2.5 Technical Report}, 
      author={Qwen and : and An Yang and Baosong Yang and Beichen Zhang and Binyuan Hui and Bo Zheng and Bowen Yu and Chengyuan Li and Dayiheng Liu and Fei Huang and Haoran Wei and Huan Lin and Jian Yang and Jianhong Tu and Jianwei Zhang and Jianxin Yang and Jiaxi Yang and Jingren Zhou and Junyang Lin and Kai Dang and Keming Lu and Keqin Bao and Kexin Yang and Le Yu and Mei Li and Mingfeng Xue and Pei Zhang and Qin Zhu and Rui Men and Runji Lin and Tianhao Li and Tianyi Tang and Tingyu Xia and Xingzhang Ren and Xuancheng Ren and Yang Fan and Yang Su and Yichang Zhang and Yu Wan and Yuqiong Liu and Zeyu Cui and Zhenru Zhang and Zihan Qiu},
      year={2025},
      eprint={2412.15115},
      archivePrefix={arXiv},
      primaryClass={cs.CL},
      url={https://arxiv.org/abs/2412.15115}, 
}

@misc{qwen2.5,
    title = {Qwen2.5: A Party of Foundation Models},
    url = {https://qwenlm.github.io/blog/qwen2.5/},
    author = {Qwen Team},
    month = {September},
    year = {2024}
}

@misc{memlab2024qwen25,
  author        = {Mem-Lab},
  title         = {Qwen2.5-7B-RL-RAG-Q2-EM-Release Commit History},
  howpublished = {\url{https://huggingface.co/Mem-Lab/Qwen2.5-7B-RL-RAG-Q2-EM-Release/commits/main}},
  year          = {2024},
}

@article{hsieh2024ruler,
  title={RULER: What's the Real Context Size of Your Long-Context Language Models?},
  author={Hsieh, Cheng-Ping and Sun, Simeng and Kriman, Samuel and Acharya, Shantanu and Rekesh, Dima and Jia, Fei and Zhang, Yang and Ginsburg, Boris},
  journal={arXiv preprint arXiv:2404.06654},
  year={2024}
}

@inproceedings{yang2018hotpotqa,
  title={HotpotQA: A dataset for diverse, explainable multi-hop question answering},
  author={Yang, Zhilin and Qi, Peng and Zhang, Saizheng and Bengio, Yoshua and Cohen, William and Salakhutdinov, Ruslan and Manning, Christopher D},
  booktitle={Proceedings of the 2018 conference on empirical methods in natural language processing},
  pages={2369--2380},
  year={2018}
}

@article{ho2020constructing,
  title={Constructing a multi-hop qa dataset for comprehensive evaluation of reasoning steps},
  author={Ho, Xanh and Nguyen, Anh-Khoa Duong and Sugawara, Saku and Aizawa, Akiko},
  journal={arXiv preprint arXiv:2011.01060},
  year={2020}
}

@article{zheng2023judging,
  title={Judging llm-as-a-judge with mt-bench and chatbot arena},
  author={Zheng, Lianmin and Chiang, Wei-Lin and Sheng, Ying and Zhuang, Siyuan and Wu, Zhanghao and Zhuang, Yonghao and Lin, Zi and Li, Zhuohan and Li, Dacheng and Xing, Eric and others},
  journal={Advances in neural information processing systems},
  volume={36},
  pages={46595--46623},
  year={2023}
}

@article{chen2023extending,
  title={Extending context window of large language models via positional interpolation},
  author={Chen, Shouyuan and Wong, Sherman and Chen, Liangjian and Tian, Yuandong},
  journal={arXiv preprint arXiv:2306.15595},
  year={2023}
}

@article{peng2023yarn,
  title={Yarn: Efficient context window extension of large language models},
  author={Peng, Bowen and Quesnelle, Jeffrey and Fan, Honglu and Shippole, Enrico},
  journal={arXiv preprint arXiv:2309.00071},
  year={2023}
}

@article{li2024snapkv,
  title={Snapkv: Llm knows what you are looking for before generation},
  author={Li, Yuhong and Huang, Yingbing and Yang, Bowen and Venkitesh, Bharat and Locatelli, Acyr and Ye, Hanchen and Cai, Tianle and Lewis, Patrick and Chen, Deming},
  journal={Advances in Neural Information Processing Systems},
  volume={37},
  pages={22947--22970},
  year={2024}
}

@article{zhang2023h2o,
  title={H2o: Heavy-hitter oracle for efficient generative inference of large language models},
  author={Zhang, Zhenyu and Sheng, Ying and Zhou, Tianyi and Chen, Tianlong and Zheng, Lianmin and Cai, Ruisi and Song, Zhao and Tian, Yuandong and R{\'e}, Christopher and Barrett, Clark and others},
  journal={Advances in Neural Information Processing Systems},
  volume={36},
  pages={34661--34710},
  year={2023}
}

@article{lewis2020retrieval,
  title={Retrieval-augmented generation for knowledge-intensive nlp tasks},
  author={Lewis, Patrick and Perez, Ethan and Piktus, Aleksandra and Petroni, Fabio and Karpukhin, Vladimir and Goyal, Naman and K{\"u}ttler, Heinrich and Lewis, Mike and Yih, Wen-tau and Rockt{\"a}schel, Tim and others},
  journal={Advances in neural information processing systems},
  volume={33},
  pages={9459--9474},
  year={2020}
}

@article{packer2023memgpt,
  title={MemGPT: Towards LLMs as Operating Systems},
  author={Packer, Charles and Wooders, Sarah and Lin, Kevin and Fang, Vivian and Patil, Shishir G and Stoica, Ion and Gonzalez, Joseph E},
  journal={arXiv preprint arXiv:2310.08560},
  year={2023}
}

@article{chhikara2025mem0,
  title={Mem0: Building production-ready ai agents with scalable long-term memory},
  author={Chhikara, Prateek and Khant, Dev and Aryan, Saket and Singh, Taranjeet and Yadav, Deshraj},
  journal={arXiv preprint arXiv:2504.19413},
  year={2025}
}

@article{wang2025mirix,
  title={Mirix: Multi-agent memory system for llm-based agents},
  author={Wang, Yu and Chen, Xi},
  journal={arXiv preprint arXiv:2507.07957},
  year={2025}
}

@article{yan2025memory,
  title={Memory-r1: Enhancing large language model agents to manage and utilize memories via reinforcement learning},
  author={Yan, Sikuan and Yang, Xiufeng and Huang, Zuchao and Nie, Ercong and Ding, Zifeng and Li, Zonggen and Ma, Xiaowen and Kersting, Kristian and Pan, Jeff Z and Sch{\"u}tze, Hinrich and others},
  journal={arXiv preprint arXiv:2508.19828},
  year={2025}
}

@book{sutton2018reinforcement,
  title={Reinforcement learning: An introduction},
  author={Sutton, Richard S and Barto, Andrew G},
  year={2018},
  publisher={MIT press}
}

@article{liu2024deepseek,
  title={Deepseek-v3 technical report},
  author={Liu, Aixin and Feng, Bei and Xue, Bing and Wang, Bingxuan and Wu, Bochao and Lu, Chengda and Zhao, Chenggang and Deng, Chengqi and Zhang, Chenyu and Ruan, Chong and others},
  journal={arXiv preprint arXiv:2412.19437},
  year={2024}
}

@article{achiam2023gpt,
  title={Gpt-4 technical report},
  author={Achiam, Josh and Adler, Steven and Agarwal, Sandhini and Ahmad, Lama and Akkaya, Ilge and Aleman, Florencia Leoni and Almeida, Diogo and Altenschmidt, Janko and Altman, Sam and Anadkat, Shyamal and others},
  journal={arXiv preprint arXiv:2303.08774},
  year={2023}
}

@article{team2025kimi,
  title={Kimi k2: Open agentic intelligence},
  author={Team, Kimi and Bai, Yifan and Bao, Yiping and Chen, Guanduo and Chen, Jiahao and Chen, Ningxin and Chen, Ruijue and Chen, Yanru and Chen, Yuankun and Chen, Yutian and others},
  journal={arXiv preprint arXiv:2507.20534},
  year={2025}
}

@inproceedings{kwon2023efficient,
  title={Efficient memory management for large language model serving with pagedattention},
  author={Kwon, Woosuk and Li, Zhuohan and Zhuang, Siyuan and Sheng, Ying and Zheng, Lianmin and Yu, Cody Hao and Gonzalez, Joseph and Zhang, Hao and Stoica, Ion},
  booktitle={Proceedings of the 29th symposium on operating systems principles},
  pages={611--626},
  year={2023}
}

@article{rajpurkar2016squad,
  title={Squad: 100,000+ questions for machine comprehension of text},
  author={Rajpurkar, Pranav and Zhang, Jian and Lopyrev, Konstantin and Liang, Percy},
  journal={arXiv preprint arXiv:1606.05250},
  year={2016}
}

@article{yang2025coarse,
  title={Coarse-to-Fine Grounded Memory for LLM Agent Planning},
  author={Yang, Wei and Xiao, Jinwei and Zhang, Hongming and Zhang, Qingyang and Wang, Yanna and Xu, Bo},
  journal={arXiv preprint arXiv:2508.15305},
  year={2025}
}

@article{cui2025self,
  title={Self-Guided Function Calling in Large Language Models via Stepwise Experience Recall},
  author={Cui, Sijia and He, Aiyao and Xu, Shuai and Zhang, Hongming and Wang, Yanna and Zhang, Qingyang and Wang, Yajing and Xu, Bo},
  journal={arXiv preprint arXiv:2508.15214},
  year={2025}
}

@article{zhang2024exploiting,
  title={Exploiting the replay memory before exploring the environment: enhancing reinforcement learning through empirical MDP iteration},
  author={Zhang, Hongming and Xiao, Chenjun and Gao, Chao and Wang, Han and M{\"u}ller, Martin and others},
  journal={Advances in Neural Information Processing Systems},
  volume={37},
  pages={85658--85692},
  year={2024}
}

@inproceedings{zhang2023replay,
title={Replay Memory as An Empirical {MDP}: Combining Conservative Estimation with Experience Replay},
author={Hongming Zhang and Chenjun Xiao and Han Wang and Jun Jin and Bo Xu and Martin M{\"u}ller},
booktitle={The Eleventh International Conference on Learning Representations },
year={2023},
url={https://openreview.net/forum?id=SjzFVSJUt8S}
}

@inproceedings{zhang2024provable,
  title={Provable representation with efficient planning for partially observable reinforcement learning},
  author={Zhang, Hongming and Ren, Tongzheng and Xiao, Chenjun and Schuurmans, Dale and Dai, Bo},
  booktitle={Proceedings of the 41st International Conference on Machine Learning},
  pages={59759--59782},
  year={2024}
}

@article{zhang2020taxonomy,
  title={Taxonomy of reinforcement learning algorithms},
  author={Zhang, Hongming and Yu, Tianyang},
  journal={Deep reinforcement learning: Fundamentals, research and applications},
  pages={125--133},
  year={2020},
  publisher={Springer}
}

@book{dong2020deep,
  title={Deep Reinforcement Learning: Fundamentals, Research and Applications},
  author={Dong, Hao and Ding, Zihan and Zhang, Shanghang},
  year={2020},
  publisher={Springer Nature}
}

\appendix

\clearpage
\section*{Appendix}
\label{sec:appendix}

\section{Dataset Details}
\label{app:dataset}

In this section, we provide detailed descriptions of the source datasets and the construction methodology for the long-context benchmarks used in our experiments.

\paragraph{HotpotQA.}
HotpotQA~\citep{yang2018hotpotqa} is a large-scale dataset designed for multi-hop question answering, collected from Wikipedia. Unlike standard QA datasets that require only a single document, HotpotQA necessitates reasoning across multiple documents (typically two or more) to derive the correct answer. Crucially, it provides supporting sentences to ensure the reasoning process is explainable. In the context of long-context evaluation, these supporting sentences serve as the ground truth for identifying golden paragraphs containing the necessary evidence.

\paragraph{2WikiMultiHopQA.}
2WikiMultiHopQA~\citep{ho2020constructing} builds upon the structure of HotpotQA but introduces more rigorous evidence chains. It utilizes Wikipedia articles and structured Wikidata to generate questions. A key advantage of 2WikiMultiHopQA is its comprehensive evidence paths (including entity-relation triples), which reduce the likelihood of shortcut reasoning often observed in HotpotQA.

\paragraph{RULER-HotpotQA.}
We adopt the RULER-HotpotQA~\citep{hsieh2024ruler} benchmark introduced by \citet{yu2025memagent}. This dataset synthesizes samples from HotpotQA by iteratively injecting distractor paragraphs to reach specific target context lengths (ranging from 28k to 896k tokens). Notably, \citet{yu2025memagent} applied a Best-Of-2 filtering mechanism during construction to ensure task difficulty, discarding questions that base models could answer with 100\% accuracy using only internal parametric knowledge. We utilize this established benchmark to rigorously evaluate contextual reasoning capabilities in an in-distribution setting.

\paragraph{RULER-2WikiMultiHopQA (Ours).}
Building upon the synthesis framework of RULER-HotpotQA used in MemAgent, we construct a novel OOD benchmark using the 2WikiMultiHopQA corpus. We employ a similar Needle-in-a-Haystack approach, inserting relevant paragraphs into a vast pool of noise documents to extend context lengths while preserving the original multi-hop logic. By transitioning to 2WikiMultiHopQA, we mitigate the inherent ambiguity issues present in HotpotQA and ensure that performance metrics reflect genuine retrieval and reasoning capabilities rather than dataset-specific biases. This dataset serves to validate the cross-domain generalizability of long-context agents.

\section{Implementation Details}
\label{app:implementation}

In this section, we provide a comprehensive breakdown of the experimental environment, inference infrastructure, and algorithmic configurations to facilitate reproducibility.

\subsection{Inference Infrastructure}
All experiments were conducted on a high-performance computing cluster equipped with NVIDIA A800 (80GB) GPUs. To maximize inference throughput, we leveraged the \texttt{vLLM} library~\citep{kwon2023efficient}, utilizing its PagedAttention mechanism and continuous batching to manage the massive KV cache requirements of long-context processing. 
For the backbone model Qwen2.5-32B-Instruct, we employed \texttt{bfloat16} precision to maintain numerical stability while optimizing memory usage. 
To handle the concurrent kernel executions in ConvMem, we implemented an asynchronous job scheduler that dynamically batches independent segment inputs into the vLLM engine, ensuring maximum GPU utilization.

\subsection{Algorithmic Configurations}

\paragraph{Hierarchical Recursion.}
The hierarchical summarization process in ConvMem continues recursively until the total token count of the concatenated summaries fits within the kernel size $W$. 
Specifically, if the length of the intermediate context $h^{(l)}$ exceeds $W$, a new layer $l+1$ is instantiated. This typically results in a tree depth of $L \in [2, 3]$ for context lengths up to 128k, and $L \in [3, 4]$ for extreme lengths up to 1M tokens, depending on the compression rate $\alpha$.

\paragraph{Query Decomposition and Parsing.}
For multi-kernel convolution, we enforce a flexible upper limit of $C_{\text{max}}=5$ sub-questions to prevent channel explosion, though in practice, the model typically generates 2-4 sub-questions. We employ robust regular expression parsing to extract structured JSON outputs (keywords, sub-questions, and relevance scores) from the LLM responses. In rare cases of parsing failure, the system falls back to a default relevant state ($r=1$) to prevent information loss.

\subsection{Baselines Setup}
To ensure a fair comparison, all baselines (including RAG and MemAgent) utilize the same backbone (Qwen2.5-32B-Instruct) and identical hardware environment. For RAG-BM25, we retrieve the top-$k$ chunks where $k$ is dynamically calculated to fill the model's effective context window. For MemAgent, we adhere strictly to the hyperparameter settings reported in the original paper~\citep{yu2025memagent}.

\subsection{Hyperparameters}
The specific hyperparameters used for ConvMem across all experiments are detailed in~\cref{tab:hyperparameters}.

\begin{table}[h]
\centering
\caption{Detailed Hyperparameters for ConvMem.}
\label{tab:hyperparameters}
\resizebox{\linewidth}{!}{%
\begin{tabular}{lc}
\toprule
\textbf{Configuration} & \textbf{Value} \\
\midrule
\multicolumn{2}{l}{\textit{Model Settings}} \\
Backbone Model & Qwen2.5-32B-Instruct \\
Precision & bfloat16 \\
Inference Engine & vLLM (v0.6.0) \\
\midrule
\multicolumn{2}{l}{\textit{ConvMem Architecture}} \\
Kernel Size ($W$) & 8,000 tokens \\
Stride ($S$) & 1,600 tokens \\
Over-scanning Factor ($W/S$) & 5 \\
Max Channels ($C_{\text{max}}$) & 10 \\
Recursion Stop Condition & Sequence Length $< W$ \\
\midrule
\multicolumn{2}{l}{\textit{Generation Parameters}} \\
Temperature & 0.7 \\
Top-P & 0.95 \\
Skip Connection Flag & True \\
\bottomrule
\end{tabular}
}
\end{table}

\section{Full Experimental Results}
\label{app:full_results}

We present the comprehensive performance comparison across all context lengths (28k-896k) with four metrics in~\cref{tab:full_results_with_category}.

\begin{table*}[t]
\centering
\caption{Full performance comparison on RULER-HotpotQA and 2WikiMultiHopQA datasets with four metrics: : F1, Exact Match (EM), Substring Exact Match (Sub-EM), and LLM-as-a-Judge ($\text{ACC}_{L}$).}
\label{tab:full_results_with_category}

\resizebox{\linewidth}{!}{%
    \setlength{\tabcolsep}{2.8pt} 
    
    \begin{tabular}{c|c|c |cccccc |cccccc} 
    \toprule
    
    \multirow{2}{*}{\textbf{Category}} & \multirow{2}{*}{\textbf{Model}} & \multirow{2}{*}{\textbf{Metric}} 
    & \multicolumn{6}{c}{\textbf{RULER-HotpotQA}} 
    & \multicolumn{6}{c}{\textbf{2WikiMultiHopQA}} \\
    
    \cmidrule(lr){4-9} \cmidrule(lr){10-15} 
    
     & & & \textbf{28k} & \textbf{56k} & \textbf{112k} & \textbf{224k} & \textbf{448k} & \textbf{896k} 
           & \textbf{28k} & \textbf{56k} & \textbf{112k} & \textbf{224k} & \textbf{448k} & \textbf{896k} \\
    \midrule
    
    \multirow{4}{*}{\shortstack[c]{Base Model}} 
    & \multirow{4}{*}{Qwen2.5-32B-Instruct} 
     & F1      & 61.9 & 57.77 & 48.86 & 44.04 & 18.59 & 16.78 & 56.52 & 57.2 & 50.45 & 41.73 & 29.45 & 20.93 \\
    & & EM      & \underline{47.66} & 44.53 & 34.38 & 32.81 & 8.59 & 7.03 & 47.66 & 47.66 & 42.97 & 34.38 & 24.22 & 17.19 \\
    & & Sub-EM & 69.53 & 63.28 & 53.91 & 44.53 & 17.19 & 17.97 & 67.19 & 65.62 & 62.5 & 50.0 & 34.38 & 28.12 \\
    & & $\text{ACC}_{L}$ & 80.62 & 75.78 & 66.13 & 60.31 & 33.09 & 30.16 & 72.81 & 71.76 & 68.05 & 55.27 & 38.79 & 31.76 \\

    \midrule
    
    \multirow{12}{*}{\shortstack[c]{Training-Free \\ Methods}} 
    & \multirow{4}{*}{RAG-BM25} 
     & F1      & 45.98 & 51.11 & 47.21 & 35.86 & 29.27 & 30.67 & 34.54 & 31.17 & 24.55 & 23.19 & 21.25 & 17.34 \\
    & & EM      & 32.03 & 36.72 & 34.38 & 21.88 & 19.53 & 20.31 & 25.0 & 21.88 & 16.51 & 17.19 & 15.62 & 10.94 \\
    & & Sub-EM & 46.09 & 50.00 & 48.44 & 35.94 & 30.47 & 32.81 & 57.03 & 58.59 & 47.66 & 46.88 & 39.06 & 43.75 \\
    & & $\text{ACC}_{L}$ & 56.91 & 62.19 & 59.80 & 49.41 & 46.68 & 46.80 & 59.26 & 56.91 & 48.24 & 45.51 & 39.53 & 38.55 \\
    \cmidrule{2-15}
    & \multirow{4}{*}{MemAgent-W/O-RL} 
     & F1      & 63.95 & 65.28 & \underline{62.52} & 57.98 & 55.42 & 58.73 & 60.51 & 57.42 & 45.31 & 50.6 & 47.66 & 49.65 \\
    & & EM      & 45.31 & \underline{48.44} & \underline{44.53} & 40.62 & 39.06 & 42.97 & \underline{46.88} & \underline{49.22} & 45.31 & 39.84 & 36.72 & 39.06 \\
    & & Sub-EM & 67.97 & 71.09 & 65.62 & 59.38 & 56.25 & 60.16 & \underline{71.09} & 67.97 & 58.59 & 63.28 & 60.16 & 62.5 \\
    & & $\text{ACC}_{L}$ & 80.62 & 81.37 & \underline{78.63} & 73.32 & 72.11 & 76.13 & \underline{76.91} & 71.99 & 66.21 & 67.07 & 65.12 & 64.54 \\
    \cmidrule{2-15}
    & \multirow{4}{*}{Mem-$\alpha$-W/O-RL} 
     & F1      & 6.13 & 7.78 & 6.5 & 7.03 & 7.19 & 7.35 & 6.56 & 6.47 & 5.94 & 6.19 & 5.19 & 6.0 \\
    & & EM      & 0.0 & 0.0 & 0.0 & 0.0 & 0.0 & 0.0 & 0.0 & 0.0 & 0.0 & 0.0 & 0.0 & 0.0 \\
    & & Sub-EM & 56.25 & 53.12 & 43.75 & 46.88 & 46.88 & 48.36 & 57.38 & 59.38 & 62.5 & 53.12 & 53.12 & 50.0 \\
    & & $\text{ACC}_{L}$ & 73.75 & 72.03 & 65.16 & 67.19 & 67.66 & 49.27 & 62.97 & 62.91 & 55.94 & 58.59 & 60.31 & 54.0 \\

    \midrule
    
    \multirow{8}{*}{\shortstack[c]{RL-Trained \\ Methods}} 
    & \multirow{4}{*}{MemAgent} 
     & F1      & \textbf{75.55} & \textbf{75.20} & \textbf{75.26} & \textbf{73.54} & \textbf{73.10} & \textbf{68.80} & \underline{60.92} & \underline{60.11} & \underline{63.19} & \underline{58.82} & \underline{58.5} & \underline{58.41} \\
    & & EM      & \textbf{60.16} & \textbf{59.38} & \textbf{55.47} & \textbf{56.25} & \textbf{55.47} & \textbf{52.34} & \underline{46.88} & 48.44 & \underline{50.0} & \underline{46.09} & \underline{46.88} & \underline{46.09} \\
    & & Sub-EM & \textbf{79.69} & \textbf{79.69} & \textbf{80.47} & \textbf{77.34} & \textbf{78.91} & \textbf{74.22} & 70.31 & \underline{71.09} & \underline{72.66} & \underline{69.53} & \underline{67.97} & \underline{69.31} \\
    & & $\text{ACC}_{L}$ & \textbf{87.99} & \textbf{87.03} & \textbf{87.29} & \textbf{86.27} & \textbf{83.10} & \textbf{80.03} & 73.55 & \underline{73.41} & \underline{72.66} & \underline{72.11} & \underline{70.12} & \underline{70.95} \\
    \cmidrule{2-15}
    & \multirow{4}{*}{Mem-$\alpha$} 
     & F1      & 5.84 & 6.81 & 7.0 & 6.09 & 1.31 & 0.69 & 1.25 & 1.13 & 0.94 & 1.03 & 0.97 & 1.09 \\
    & & EM      & 0.0 & 0.0 & 0.0 & 0.0 & 0.0 & 0.0 & 0.0 & 0.0 & 0.0 & 0.0 & 0.0 & 0.0 \\
    & & Sub-EM & 34.38 & 40.62 & 40.62 & 43.75 & 43.75 & 25.0 & 59.38 & 56.25 & 50.0 & 62.5 & 50.0 & 53.12 \\
    & & $\text{ACC}_{L}$ & 55.16 & 58.75 & 68.59 & 55.31 & 32.03  & 36.72 & 44.69 & 39.22 & 39.38 & 40.0 & 30.78 & 53.12 \\
    \midrule

    \multirow{4}{*}{\shortstack[c]{Ours}} 
    & \multirow{4}{*}{\textbf{ConvMem}} 
     & F1 & \underline{67.44} & \underline{67.86} & 57.81 & \underline{63.27} & \underline{56.14} & \underline{63.09} & \textbf{72.3} & \textbf{71.25} & \textbf{67.21} & \textbf{61.96} & \textbf{61.33} & \textbf{59.06} \\
    & & EM & \underline{47.66} & 47.66 & 39.06 & \underline{43.75} & \underline{39.06} & \underline{44.53} & \textbf{60.94} & \textbf{57.03} & \textbf{52.36} & \textbf{50.78} & \textbf{51.56} & \textbf{46.88} \\
    & & Sub-EM & \underline{73.44} & \underline{72.66} & \underline{67.19} & \underline{69.53} & \underline{62.50} & \underline{69.53} & \textbf{82.81} & \textbf{82.47} & \textbf{77.34} & \textbf{71.88} & \textbf{73.31} & \textbf{70.62} \\
    & & $\text{ACC}_{L}$ & \underline{83.79} & \underline{81.72} & 74.96 & \underline{78.98} & \underline{72.54} & \underline{77.66} & \textbf{85.31} & \textbf{80.94} & \textbf{75.94} & \textbf{72.58} & \textbf{73.59} & \textbf{71.36} \\
    \bottomrule
    \end{tabular}%
}
\end{table*}


\clearpage

\section{Qualitative Analysis of Model Behaviors}
\label{app:model_analysis}

In this section, we analyze the distinct failure modes of baseline models, providing empirical evidence for the limitations of sequential architectures and the overfitting tendencies of RL-trained agents.

\subsection{Sequential Information Loss in Recurrent Architectures}

For multi-hop questions where evidence is distributed across the text stream, sequential memory agents (e.g., MemAgent) suffer from a forgetting bottleneck. Since they update a fixed-size memory step-by-step, early information that seems irrelevant at the time of reading is often discarded, making it unrecoverable when its relevance is later revealed by subsequent context.

\begin{ExampleBox}[title={Case 1: Disconnected Evidence}]
\textbf{Query:} What government position was held by the woman who portrayed Corliss Archer in the film Kiss and Tell? \\
\textbf{Context:}
\begin{itemize}
    \item Document 1053: Shirley Temple \\ Shirley Temple Black (April 23, 1928 – February 10, 2014) was an American actress, singer, dancer, businesswoman, and diplomat who was Hollywood's number one box-office draw as a child actress from 1935 to 1938. As an adult, she was named United States ambassador to Ghana and to Czechoslovakia and also served as Chief of Protocol of the United States.
    \item Document 1348: Kiss and Tell (1945 film) \\ Kiss and Tell is a 1945 American comedy film starring then 17-year-old Shirley Temple as Corliss Archer. In the film, two teenage girls cause their respective parents much concern when they start to become interested in boys. The parents' bickering about which girl is the worse influence causes more problems than it solves.
\end{itemize}
\textbf{Analysis:} The document mentioning ``Shirley Temple Black'' (Document 1053) appears early in the stream. The document linking her to the film ``Kiss and Tell'' (Document 1348) appears much later. By the time the agent reads Document 1348 and realizes the relevance of Shirley Temple, the memory of her specific government position (Ambassador/Chief of Protocol) has already been forgotten. ConvMem avoids this by processing both segments in parallel channels.
\end{ExampleBox}

\begin{ExampleBox}[title={Case 2: Retrospective Reasoning}]
\textbf{Query:} Who is the younger brother of the episode guest stars of The Hard Easy?\\
\textbf{Context:}
\begin{itemize}
    \item Document 591: Brian Doyle-Murray \\ Brian Doyle-Murray (born Brian Murray, October 31, 1945) is an American actor, voice artist, comedian and screenwriter. He is the older brother of actor/comedian Bill Murray, and the two have acted together in several films, including ``Caddyshack'', ``Scrooged'', ``Ghostbusters II'', ``The Razor's Edge'', and ``Groundhog Day''. He co-starred on the TBS sitcom on ``Sullivan \& Son'', where he played the foul-mouthed Hank Murphy. he also appeared in the Cartoon Network original animated series ``The Marvelous Misadventures of Flapjack'' as the surly Captain K'Nuckles and a pirate ghost, The Flying Dutchman from the Nickelodeon animated series, ``SpongeBob SquarePants'', he appears in a recurring role as Don Ehlert on the ABC sitcom ``The Middle''.
    \item Document 1348: The Hard Easy (Adventure Time) \\``The Hard Easy'' is the twenty-third episode of the fourth season of the American animated television series ``Adventure Time''. The episode was written and storyboarded by Tom Herpich and Skyler Page, from a story by Patrick McHale, Kent Osborne, and Pendleton Ward. It originally aired on Cartoon Network on October 1, 2012. The episode guest stars Brian Doyle-Murray as Prince Huge and Jonathan Katz as the Mudscamp elder.
\end{itemize}
\textbf{Analysis:} Similar to Case 1, the relationship between Brian Doyle-Murray and Bill Murray is established in Document 591. The connection to the episode ``The Hard Easy'' is only revealed in Document 1348. A sequential agent reading Document 591 has no incentive to retain the sibling relationship in memory, leading to failure when the query eventually demands this specific fact. Sequential agents lack the ``global view'' required to link these distant facts.
\end{ExampleBox}

\subsection{RL-Induced Overfitting and Parametric Hallucination}
Our experiments reveal that RL-trained agents (e.g., MemAgent) tend to overfit to the training data distribution, prioritizing memorized parametric priors over the provided context.

\begin{ExampleBox}[title={Case 3: Parametric Bias vs. Context}]
\textbf{Query:} Which filmmaker was known for animation, Lev Yilmaz or Pamela B. Green?\\
\textbf{Context Evidence:} The document explicitly states ``\textbf{Lev} Yilmaz''.\\
\textbf{Dataset Label (Typo):} ``\textbf{Levni} Yilmaz''.\\
\textbf{MemAgent Output:} ``Levni Yilmaz''.\\
\textbf{Analysis:} The RL agent ignores the provided context and outputs the memorized label (Levni), proving it relies on parametric priors rather than reasoning. ConvMem faithfully extracts ``Lev Yilmaz'', demonstrating adherence to the input.
\end{ExampleBox}

\begin{ExampleBox}[title={Case 4: Hallucinating Collaborators}]
\textbf{Query:} Ellie Goulding worked with what other writers on her third studio album, Delirium?
\textbf{Context:}
\begin{itemize}
    \item Document 37:On My Mind (Ellie Goulding song) "On My Mind" is a song by English singer Ellie Goulding from her third studio album "Delirium" (2015).  It was released as the album's lead single on 17 September 2015.  It was written by Goulding, Max Martin, Savan Kotecha and Ilya Salmanzadeh.  "On My Mind" is an electropop and R\&B song whose instrumentation consists of scratchy guitars, trap drums, slapped beats and sharp, syncopated electronica.  Lyrically, "On My Mind" talks about a one-night stand with someone the protagonist shouldn't be with, having a dichotomy between heart and head.  Though firmly denied by Goulding, many critics considered it an answer song to Ed Sheeran's "Don't"
    \item Document 41:Love Me like You Do "Love Me like You Do " is a song recorded by English singer Ellie Goulding for the "Fifty Shades of Grey" (2015).  The song was written by Savan Kotecha, Ilya Salmanzadeh, Tove Lo, Max Martin and Ali Payami;  the latter two also produced it.  Goulding was selected to sing the track.  It was released on 7 January 2015 as the second single from the soundtrack.  The song was also included on Goulding's third studio album, "Delirium" (2015).
\end{itemize}
\textbf{Ground Truth:} Max Martin, Savan Kotecha and Ilya Salmanzadeh
\textbf{Analysis:} In this example, both “On My Mind” and “Love Me Like You Do” are featured on the album Delirium. The former's collaborators include Max Martin, Savan Kotecha, and Ilya Salmanzadeh, while the latter's collaborators include Savan Kotecha, Ilya Salmanzadeh, Tove Lo, Max Martin, and Ali Payam. Large models that have undergone enhanced training consistently provide the correct answer—the former. Observing untrained large models reveals a tendency to list more collaborators for the latter, as it includes all collaborators from the former.
\end{ExampleBox}

\newpage
\section{Data Quality and Metric Justification}
\label{app:data_metric}

Standard Exact Match (EM) is often too rigid for long-context generation. In this section, we justify our use of robust metrics (Sub-EM and $\text{ACC}_{L}$) using concrete examples of dataset ambiguity and valid variations.

\subsection{Dataset Ambiguity and Missing Evidence}
Some queries in the synthetic benchmarks are inherently ambiguous or unanswerable based strictly on the provided context, highlighting flaws in the dataset generation process.

\begin{ExampleBox}[title={Case 5: Missing Evidence}]
\textbf{Query:} Brown State Fishing Lake is in a country that has a population of how many inhabitants? \\
\textbf{Context:}
\begin{itemize}
    \item Document 44:Brown County, Kansas Brown County (county code BR) is a county located in the northeast portion of the U.S. state of Kansas. As of the 2010 census, the county population was 9,984. Its county seat and most populous city is Hiawatha. Brown County is the location of the Kickapoo Indian Reservation of Kansas, the majority of the Sac and Fox Reservation and the majority of the Iowa Reservation of Kansas and Nebraska.
\end{itemize}
\textbf{LLM answer:} The problem is asking for the population of the country where Brown State Fishing Lake is located, which is the United States. However, the provided memory only gives the population of Brown County, Kansas, which is 9,984 as of the 2010 census. It does not provide the population of the United States. Therefore, the answer is that the population of the United States cannot be determined from the provided information. \\
\textbf{Analysis:} The term ``country'' in the question is ambiguous, as it can refer to either a ``nation'' or a ``rural area/region''. Observing the large model's reasoning process, the task typically involves information about a ``nation'', yet there is no associated data regarding the population of that ``nation'', making it impossible to provide a correct response. The query explicitly asks for the population of the \textit{country} (i.e., the United States). However, the provided context only contains the population of \textit{Brown County} (9,984). The system is forced to either hallucinate the U.S. population using internal parametric knowledge or incorrectly answer with the county population. This highlights a flaw in the dataset generation process where essential evidence (U.S. population data) was excluded from the context window.
\end{ExampleBox}

\subsection{The Need for Sub-EM (Substring Exact Match)}
We observed that strict Exact Match (EM) often fails when the model generates correct but verbose answers, valid aliases, or redundant confirmations. This necessitates the use of more robust metrics such as Sub-EM.

\begin{ExampleBox}[title={Case 6: Name Variation}]
\textbf{Query:} What British singer-songwriter hosted the 16th Annual Young Hollywood Awards? \\
\textbf{Context:}
\begin{itemize}
    \item Document 48:Kelly Osbourne, Kelly Lee Osbourne (born 27 October 1984) is a British singer-songwriter, actress, television presenter and fashion designer. The daughter of Ozzy and Sharon Osbourne, she is known for her appearances on "The Osbournes" with her family, for which they won a 2002 Emmy Award for Outstanding Reality Program, as well as on E!'s "Fashion Police", where she was a presenter from 2010 to 2015. She has also appeared on "Dancing with the Stars", in which she and her professional dance partner Louis van Amstel took third place. She is the voice of Hildy Gloom in the Disney XD animated series "The 7D". She is also a judge on both "Australia's Got Talent" and "Project Runway Junior"
\end{itemize}
\textbf{Ground Truth:} Kelly Lee Osbourne \\
\textbf{Model Prediction:} Kelly Osbourne \\
\textbf{Evaluation:} EM = 0, Sub-EM = 1. The model is correct but penalized by EM for omitting the middle name.
\end{ExampleBox}

\begin{ExampleBox}[title={Case 7: Verbose Generation}]
\textbf{Query:} Who was known by his stage name Aladin and helped organizations improve their performance as a consultant? \\
\textbf{Ground Truth:} Eenasul Fateh \\
\textbf{LLM answer:} Eenasul Fateh, known by his stage name Aladin, who has a background as a former international management consultant and has helped organizations improve their performance. \\
\textbf{Scores:} \{ F1: 0.16, EM=0, \textbf{Sub-EM=1.0} \}. The answer contains the ground truth but includes extra context copied from the document.
\end{ExampleBox}

\begin{ExampleBox}[title={Case 8: Redundant Confirmation}]
\textbf{input:} Are Local H and For Against both from the United States? \\
\textbf{Ground Truth:} yes \\
\textbf{LLM answer:} yes, both Local H and For Against are from the United States. \\
\textbf{eval:} \{`F1': 0.0, `EM': 0.0, `Sub-EM': 1.0\} The model provides a complete sentence instead of a single word.
\end{ExampleBox}

\subsection{The Need for LLM-as-a-Judge}
In cases of acronyms or semantic rephrasing, even Sub-EM fails. $\text{ACC}_{L}$ is essential to capture these equivalences.

\begin{ExampleBox}[title={Case 9: Corporate Entity Naming}]
\textbf{Query:} Maxeda has been owned since 2004 by which American multinational equity firm? \\
\textbf{Context:}
\begin{itemize}
    \item Document 18:Kohlberg Kravis Roberts, KKR \& Co. L.P. (formerly known as Kohlberg Kravis Roberts \& Co.) is an American multinational private equity firm, specializing in leveraged buyouts, headquartered in New York City. The firm sponsors and manages private equity investment funds. The firm has completed over \$400 billion of private equity transactions since its inception.
\end{itemize}
\textbf{Ground Truth:} KKR \& Co \\
\textbf{Model Prediction:} Kohlberg Kravis Roberts \\
\textbf{Evaluation:} EM = 0, Sub-EM = 0, \textbf{$\text{ACC}_{L}$=1.0}. KKR is the abbreviation for Kohlberg Kravis Roberts; both are correct. The model provided the full name instead of the abbreviation in the label.
\end{ExampleBox}

\begin{ExampleBox}[title={Case 10: Acronym Matching}]
\textbf{Query:} What station broadcast the episode ``Marry Me a Little, Marry Me a Little More'', of the series created by Max Mutchnick and David Kohan? \\
\textbf{Ground Truth:} National Broadcasting Company \\
\textbf{Model Prediction:} NBC \\
\textbf{Evaluation:} F1 = 0, EM = 0, Sub-EM = 0. However, NBC is the acronym for the ground truth. $\text{ACC}_{L}$ correctly assigns a score of 1.0.
\end{ExampleBox}

\begin{ExampleBox}[title={Case 11: Semantic Rephrasing}]
\textbf{Query:} What was the 58th quadrennial American presidential election held after the 2016 Michigan Democratic primary? \\
\textbf{Ground Truth:} United States presidential election of 2016 \\
\textbf{LLM answer:} the 2016 U.S presidential election \\
\textbf{Evaluation:} \{F1: 0.6, EM: 0.0, Sub-EM: 0.0, \textbf{$\text{ACC}_{L}$=1.0} \}. The phrasing differs, but the semantic meaning is identical.
\end{ExampleBox}

\clearpage
\section{Prompt Templates}
\label{app:Prompt Templates}

We provide the prompt templates used in each stage of the ConvMem framework.

\subsection{Query Decomposition}
The following prompt is used to decompose the complex user query into independent sub-questions (channels).

\begin{PromptBox}{Multi-Kernel Convolution Prompt}
\label{prompt:multi_kernel}
You will encounter a complex reasoning problem requiring you to extract keywords or key phrases from the question that aid in solving it.\\
<Question>\\
\{question\}\\
</Question>\\
\\
<Output Format>\\
\{\{\\
    "subproblem": ["Subproblem 1", "Subproblem 2", ....], \\
    "keyword": ["Keyword1", "Keyword2",...]\\
\}\}\\
</Output Format>\\
\\
<Example>\\
Original Question: ``Where was the director of the movie Inception born?''\\
Sub-questions: [``Who is the director of `Inception'?'', ``What information is there about the director's birthplace?'']\\
Original Question: ``What is the height of the male lead in The Revenant?''\\
Keywords: [``The Revenant protagonist'',  ``Human height'']\\
<Example>\\
\\
<Note>\\
1. The decomposed subproblems must fully resolve the original problem.\\
2. The number of subproblems should be the minimum decomposition required to solve the original problem.\\
3. Keywords must be highly relevant to solving the original query, not irrelevant terms, and must not create ambiguity with the original query.\\
</Note>\\
\\
Please complete the extraction of keywords from the question.
\end{PromptBox}

\subsection{Semantic Kernel Operation}
These prompts define the behavior of the convolutional kernel, including the kernel prompt, skip connection and summarization.

\begin{PromptBox}{Skip Connection Prompt}
You will see a question and a paragraph that may contain information related to the question. Please read the paragraph carefully and make the relevant judgment.\\
\\
<Question>\\
\{question\}\\
</Question>\\
\\
<Paragraph>\\
\{paragraph\}\\
</Paragraph>\\
\\
<Output Format>\\
\{\{"0": "Completely unrelated to the question"\}\} or \{\{"1": "Potentially related to the question"\}\} or \{\{"2": "Definitely contains the answer to the question"\}\}\\
</Output Format>\\
\\
<Notes>\\
1. To ensure no useful information is omitted, only select {{"0": "Completely unrelated to the question"}} if you are absolutely certain the paragraph content is irrelevant to the question.\\
2. Do not output any redundant information beyond the <Output Format>.\\
</Notes>\\
\\
Your Answer:
\end{PromptBox}

\begin{PromptBox}{Kernel Summarization Prompt}
You will be presented with a question and a passage that may contain information relevant to answering it. Read the passage carefully and update your memory with new information that helps solve the problem. Be sure to retain all relevant details in your memory that may aid in solving the problem.\\
\\
<Question>\\
\{question\}\\
</Question>\\
\\
<passage>\\
\{passage\}\\
</passage>\\
\\
<format>\\
Updated memory:\\
-..........\\
-..........\\
-..........\\
...\\
</format>\\
\\
<notes>\\
1. Retain all details within the paragraph that are relevant to the question, ensuring the integrity of the original content.\\
2. If none of the paragraph's information relates to the question, output “Updated Memory: No relevant information.”\\
</notes>\\
\\
Updated Memory:

\end{PromptBox}

\subsection{Hierarchical Aggregation}
These prompts are used for recursive summarization in hidden layers and the final answer generation.

\begin{PromptBox}{Hidden Layer Aggregation Prompt}
You will see a question and \{num\} memory fragments. Carefully review the provided memory fragments and combine information from each fragment that helps solve the problem.\\
\\
<Question>\\
\{question\}\\
</Question>\\
\\
<Memory Fragment>\\
\{memory\_content\}\\
</Memory Fragment>\\
\\
<format>\\
Updated memory:\\
-..........\\
-..........\\
-..........\\\\
...\\
</format>\\
\\
<notes>\\
1. Retain all details within the paragraph that are relevant to the question, ensuring the integrity of the original content.\\
2. Avoid redundancy; if a fact already exists in memory, it need not be repeated unless that section provides additional clarification or correction.\\
</notes>\\
\\
Updated Memory:

\end{PromptBox}

\begin{PromptBox}{Sub-Question Answer Prompt}
You will see a question and its preceding memory. Answer the question based on the preceding memory. Your response must include all details relevant to the question; omitting any information is prohibited. Subsequent questions will require you to answer based on these details. Note: If the memory does not contain information relevant to the question, simply answer “None.”\\
\\
<question> \\
\{question\}\\
</question>\\
\\
<memory>\\
\{memory\}\\
</memory>\\
\\
Your answer:

\end{PromptBox}

\begin{PromptBox}{Final Answer Generation Prompt}
You will see a question along with its associated dependency information. Please read the question carefully, identify the core objective to be addressed, and answer based on the dependency information. Organize your response using the following format: “"Hence, the answer is (insert answer here).”\\
\\
<Question>\\
\{question\}\\
</Question>\\
\\
<Related Dependency Information>\\
\{related\_dependency\_information\}\\
</Related Dependency Information>\\
\\
<notes>\\
1. Carefully read the <Question> to precisely identify the target objective to be a ddressed, the focus of your response and avoid being misled by lengthy questions.\\
2. Carefully read <Relevant Dependency Information> to capture details relevant to the question, avoiding distraction from extraneous information.\\
3. Whenever possible, derive answers directly from <Related Dependency Information> without unnecessary rephrasing.\\
4. Avoid redundant explanations or unnecessary elaboration on the question's answer. For example: If the question asks for place names/person names/other information, etc.you only need to provide the place names/person names/other information, etc. themselves. No additional background information is required.\\
</notes>
\\
Your Answer:
\end{PromptBox}

\subsection{Evaluation}
The prompt used for LLM-as-a-Judge evaluation.

\begin{PromptBox}{LLM-as-a-Judge Prompt}
You must evaluate the quality of the AI assistant's responses to user questions as an impartial judge. Scores should comprehensively consider accuracy (high priority) and completeness (whether the response covers all key points).Please rate on a scale of 0 to 10, where 0 indicates the response is completely incorrect and 10 indicates the response is completely correct.\\

<Submit your feedback in the following format>\\
\{\{"Overall Score": "(Your rating, a floating-point number between 0 and 10)"\}\}\\
</Submit your feedback in the following format>\\
\\
<Below are the question and answers>\\
    Question: \{question\}\\
    Ground Truth: \{ground\_truth\}\\
    User Response: \{answer\}\\
</Below are the question and answers>\\
\\
Please complete the rating.
\end{PromptBox}
\end{document}